\documentclass[10pt,twocolumn,letterpaper]{article}

\usepackage[pagenumbers]{cvpr} 

\definecolor{cvprblue}{rgb}{0.4,0.49,0.74}
\usepackage[pagebackref,breaklinks,colorlinks,allcolors=cvprblue]{hyperref}

\usepackage{bbding}
\usepackage{subcaption}
\usepackage{mathrsfs}
\usepackage[ruled,vlined]{algorithm2e}
\usepackage{xspace}
\usepackage{algpseudocode}
\usepackage{amsmath}
\usepackage{array}
\usepackage{xcolor}
\usepackage{fontawesome5}
\usepackage{arydshln}
\usepackage{booktabs}  
\usepackage{multirow}  
\def\paperID{*****} 
\def\confName{CVPR}
\def\confYear{2025}
\title{Fence Theorem: Towards Dual-Objective Semantic-Structure Isolation in Preprocessing Phase for 3D Anomaly Detection}
\author{Hanzhe Liang\textsuperscript{*}\\
{\tt\small College of Computer Science and Software Engineering, Shenzhen University}\\
{\tt\small Shenzhen Audencia Financial Technology Institute, Shenzhen University}\\
{\tt\small 2023362051@email.szu.edu.cn}
\and
Jie Zhou\\
{\tt\small National Engineering Laboratory for Big Data System Computing Technology, Shenzhen University}\\
{\tt\small jie\_jpu@163.com}
\and
Xuanxin Chen\\
{\tt\small Faculty of Education, Shenzhen University}\\
{\tt\small 2022352040@email.szu.edu.cn}
\and
Tao Dai\\
{\tt\small College of Computer Science and Software Engineering, Shenzhen University}\\
{\tt\small daitao@szu.edu.cn}
\and
Jinbao Wang\textsuperscript{\dag}\\
{\tt\small National Engineering Laboratory for Big Data System Computing Technology, Shenzhen University}\\
{\tt\small Guangdong Provincial Key Laboratory of Intelligent Information Processing}\\
{\tt\small wangjb@szu.edu.cn}
\and
Can Gao\textsuperscript{\dag}\\
{\tt\small College of Computer Science and Software Engineering, Shenzhen University}\\
{\tt\small davidgao@szu.edu.cn}\\\\
\textcolor{blue}{\textbf{The Full Mathematical Proof Will Be Published in The Full Version}}
}

\begin{document}
\maketitle
\begin{abstract}
3D anomaly detection (AD) is prominent but difficult due to lacking a unified theoretical foundation for preprocessing design.
 We establish the \textbf{Fence Theorem}, formalizing preprocessing as a dual-objective semantic isolator: (1) mitigating cross-semantic interference to the greatest extent feasible and (2) confining anomaly judgements to aligned semantic spaces wherever viable, thereby establishing intra-semantic comparability. Any preprocessing approach achieves this goal through a two-stage process of Semantic-Division and Spatial-Constraints stage. Through systematic deconstruction, we theoretically and experimentally subsume existing preprocessing methods under this theorem via tripartite evidence: qualitative analyses, quantitative studies, and mathematical proofs. Guided by the Fence Theorem, we implement Patch3D, consisting of Patch-Cutting and Patch-Matching modules, to segment semantic spaces and consolidate similar ones while independently modeling normal features within each space. Experiments on Anomaly-ShapeNet and Real3D-AD with different settings demonstrate that progressively finer-grained semantic alignment in preprocessing directly enhances point-level AD accuracy, providing inverse validation of the theorem's causal logic.
\end{abstract}
\section{Introduction}
\label{sec:intro}
\begin{figure}[th] 
    \centering 
    \includegraphics[width=0.9\linewidth]{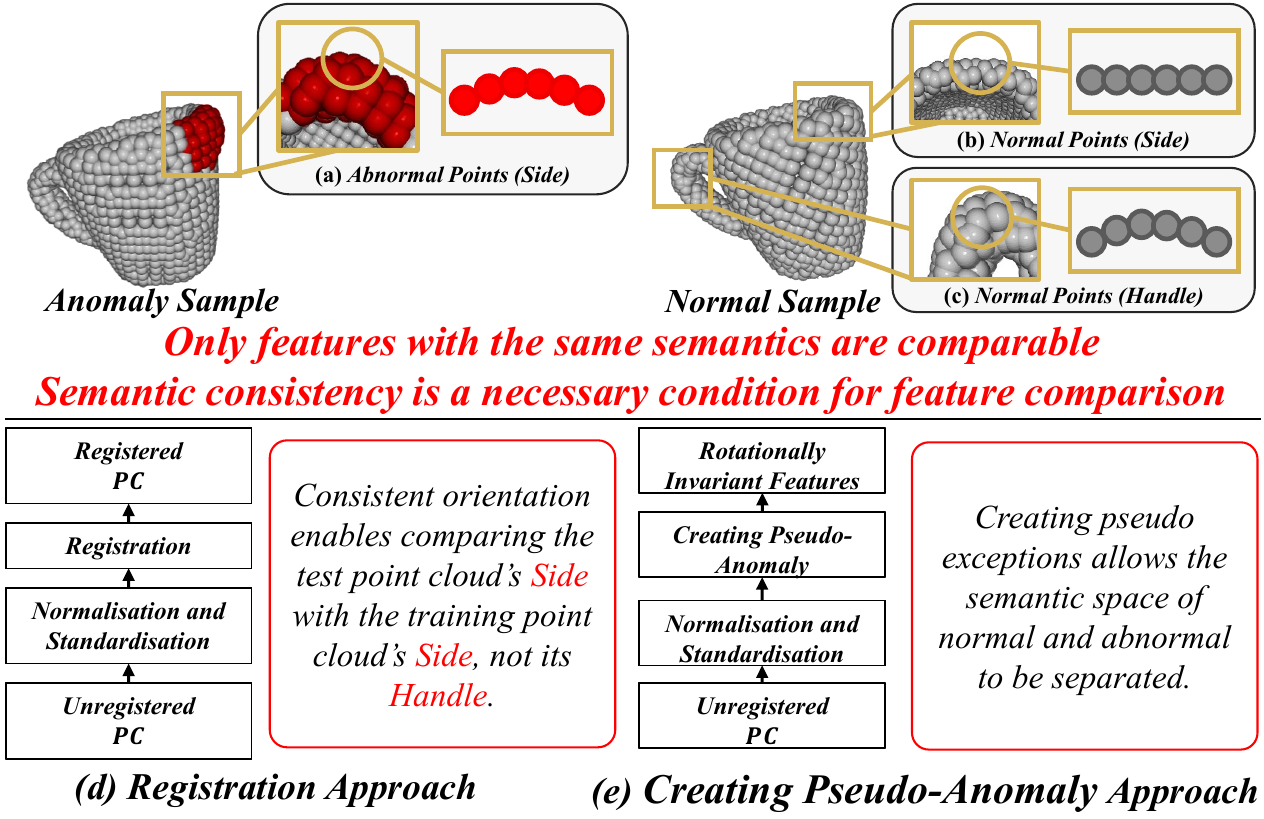} 
    \caption{\textbf{Visualisation of interference between structures.} (a) details an anomaly in the Cup wall characterized by a significant curvature. In contrast, (b) presents a normal, smooth curve on the Cup wall. (c) showcases a normal curve with a large curvature on the Cup handle, resembling the anomaly depicted in (a). Utilizing a memory bank to model the entire point cloud could result in incorrectly identifying the anomaly in (a) as normal due to the similarities with (c). Registration in (d) ensures structural similarity, while (e) ensures feature comparability via rotation-invariant embedding.
    }
    \label{motivation}
\end{figure} 
3D anomaly detection has become a hot research topic in recent years, but it has not yet been effectively explored~\cite{Mvtec3DAD,Real3D-AD,Review1,Patchcore}. The existing methods are mainly classified into feature-reconstruction and feature-embedding approach.

The feature-reconstruction approach uses the key mapping function $\mathscr{F}_1: F_{ori} \rightarrow F_{rec}$ that maps the original feature $F_{ori}$ to the reconstructed feature $F_{rec}$ to provide the model with two abilities during the training phrase: 1) The ability to regenerate normal features through the normal feature $F_{nor}$ regeneration mapping function $\mathscr{F}_2: F_{nor}\rightarrow F_{nor}$. 2) The ability to restore abnormal features $F_{abn}$ to normal features $F_{nor}$ through the abnormal feature restoration mapping function $\mathscr{F}_3: F_{abn}\rightarrow F_{nor}$ by using the preprocessing method of pseudo-anomaly generation. In the inference stage, the point cloud is mapped by $\mathscr{F}_1$, compressing its anomalous features into the normal feature distribution, with anomalies identified by $|F_{ori} - F_{rec}|$~\cite{R3DAD,IMRNet,PO3AD,Shape-Guided}. The effectiveness of the this approach depends on the veracity of the anomaly creation with the preprocessing.

The feature-embedding approach constructs a high-dimensional feature distribution $G$ from the normal feature set after being pre-processed $F_{\text{nor}} = \{ f_1, f_2, \dots, f_N \}$, where $G$ is modeled as a probability distribution $G = P(f \mid f \in F_{\text{nor}})$ during the training phase, representing the probability density function of normal features in the high-dimensional space. In the evaluation phase, for a test feature $f_{\text{eva}}$, its likelihood $P(f_{\text{eva}} \mid G) $ is computed. If $ P(f_{\text{eva}} \mid G) < \tau $, the feature $f_{\text{eva}}$ is classified as abnormal, where $\tau$ is a predefined threshold~\cite{Patchcore,Real3D-AD,Group3AD,ISMP,M3DM,BTF,AST,long2024revisitingmultimodalfusion3d,uniad}. This approach relies on different preprocessing methods, such as registration and mapping to rotation-invariant spaces.

Feature reconstruction methods and feature embedding methods use a variety of preprocessing approaches such as normalisation, standardization, pseudo-anomaly generation~\cite{R3DAD,IMRNet,PO3AD,xu2025surveyindustrialanomaliessynthesis}, and registration~\cite{ISMP,Real3D-AD,Group3AD}, and the plausibility of these approaches is crucial for 3D anomaly detection. However, the current approaches to preprocessing of individual models are still limited to their independent models, and their common purpose and mechanism of action are not yet clear.

This paper established the \textbf{Fence Theorem} to formalize preprocessing as a dual-objective semantic isolator for problems related to interpretability: mitigating cross-semantic interference and confining anomaly judgments to aligned semantic spaces. We generalized existing preprocessing approaches through qualitative analysis, quantitative verification, and mathematical proof. Guided by the theorem, we developed Patch3D, which includes Patch-Cutting and Patch-Matching modules, to decouple the semantic space and model normal features independently in each space. Experiments in Anomaly-ShapeNet and Real3D-AD showed that refined semantic alignment in preprocessing improves point-level anomaly detection accuracy and validates the theorem's logic. We make the following contributions:
\begin{itemize}
    \item We develop \textbf{Fence Theorem} to formalise preprocessing in 3D-AD as a dual-objective semantic isolator: mitigating cross-semantic interference and restricting anomaly judgments to aligned semantic spaces.
    
    \item  We categorise previously existing approaches under our Fence Theorem through empirical quantitative analysis, quantitative verification and mathematical proofs, so that the existing approaches have common purpose and mechanism of action.

    \item To support our theorem, we introduce an effective Patch3D method for modeling the distribution of normal features for anomaly detection, including the Patch-Cutting process for semantic segmentation and Patch-Matching for feature alignment. The growing trend in anomaly detection performance presented by a large number of qualitative and quantitative results for existing approaches and Patch3D's Real3D-AD and Anomaly-ShapeNet together prove the Fence Theorem.
\end{itemize}
\section{Related Work}
Recent advancements in 3D anomaly detection techniques have been instrumental in enhancing efficiency and accuracy. These techniques can be broadly categorized into two approaches~\cite{R3DAD,Review1,csad}: feature-reconstruction and feature-embedding. 

\noindent \textbf{Feature-Reconstruction Approaches.}
The feature reconstruction method is capable of detecting anomalies by measuring the difference between the origin and reconstruction data~\cite{easynet}. IMRNet~\cite{IMRNet} preprocesses the point cloud by random masking and complements the masked regions using a Masked Reconstruction Network. This approach allows the model to learn the feature representation of a normal point cloud. R3D-AD~\cite{R3DAD} proposes a diffusion model-based feature reconstruction approach, which firstly creates anomalies by a preprocessing approach and completely masks the point cloud during the diffusion process, and then gradually reconstructs the corresponding normal point cloud. The reconstruction of all points is challenging for the model; PO3AD~\cite{PO3AD} enhances the process of anomaly detection by generating pseudo-anomalies preprocessing and predicting point-level offsets, thereby ensuring the model's concentration on anomalous regions. The shape-guided approach utilises local reconstruction differences in the SDF modelling to detect anomalies. Furthermore, SplatPose~\cite{splat} and SplatPose+\cite{splatt} are based on 3DGS and achieve faster training and real-time inference by optimising 3D point cloud parameters for scene reconstruction and detecting anomalies by the difference between before and after reconstruction.

\noindent \textbf{Feature-Embedding Approaches.}
The feature embedding approach detects anomalies by comparing the features to be measured with the embedded features~\cite{tsnet}. BTF~\cite{BTF} provides an effective solution for 3D anomaly detection by combining 3D shape features and colour features to form a hybrid modal representation. M3DM~\cite{M3DM} achieves better anomaly detection through hybrid multimodal feature embedding with contrastive learning. Reg3D-AD~\cite{Real3D-AD} uses a pre-processing approach of point cloud registration by coordinates and PointMAE~\cite{PointMAE,PT} bipartite branching network to embed features. While AST~\cite{AST} uses an asymmetric student-teacher network structure using normalised streams to optimise feature embedding through positional coding and foreground masks. CPMF~\cite{CPMF} improves feature representation accuracy by external complementary pseudo-multimodal features that enable the point cloud to learn more global information. Group3AD~\cite{Group3AD}, on the other hand, optimises feature embedding in the high-resolution point cloud being pre-processed by the registration by using group-level feature contrastive learning that improves the ISMP~\cite{CPMF} optimises feature quality and alignment accuracy by extracting global features from the internal structure of the point cloud being registered and combining them with local features. In addition, large language models open up new possibilities for anomaly detection as well~\cite{ZeroShot,3DLLM,PointAD}. These approaches achieve good feature embedding through multiple preprocessing methods, offering the possibility of better anomaly detection.
\section{Fence Theorem}
The various existing preprocessing methods lack a unified theoretical foundation for preprocessing design. In this section, we present our unified theory, the Fence Theorem, established through three dimensions: qualitative research, quantitative analysis, and mathematical proof, with empirical analyses provided in subsequent sections. More complementary theorems will be reported in the \textit{supplementary material}.

\begin{figure*}[t] 
    \centering 
    \includegraphics[width=1\textwidth]{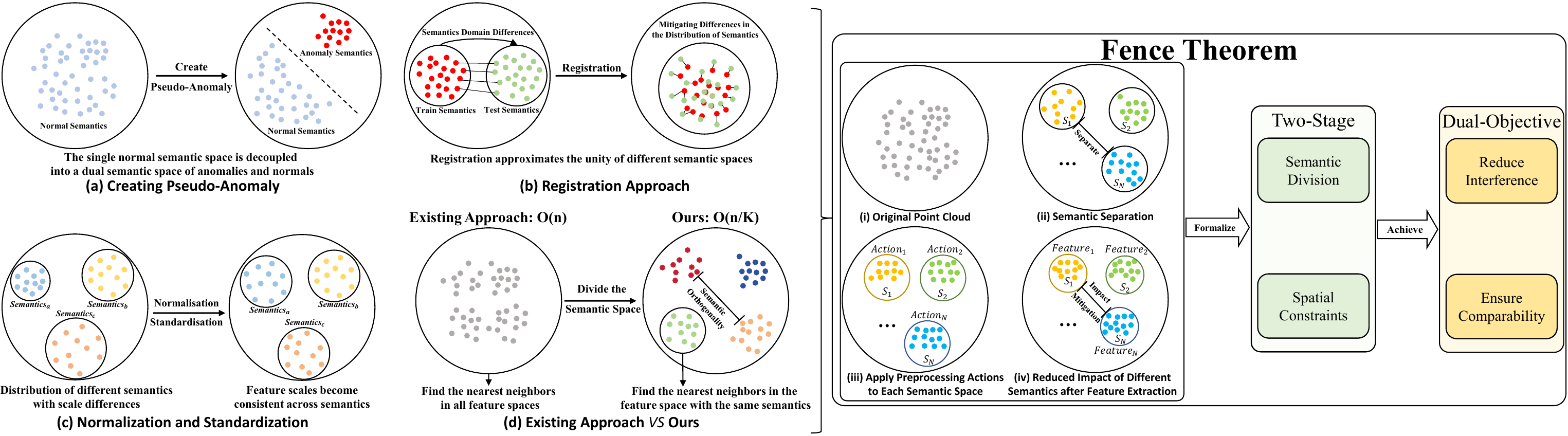} 
    \caption{Visualisation of the Fence Theorem. \label{R4}}
\end{figure*}

\subsection{Definition}
 To describe our Fence Theorem precisely, we first specify the notation. 
(1) The point cloud $\mathcal{P}$\text{=}$\{p_i\}_{i\text{=}1}^n$
 , where each point $p_i$=$(x_i, y_i, z_i, s_i)\in\mathbb{R}^3 \times \mathcal{L}$,  and $\mathcal{L} \text{=} \{1, 2, \ldots, n\}$, means that each point has its own defined semantics $s_i$.
(2) The preprocessing operation $\mathcal{A}$ is defined as a family of semantic-specific transformations $\mathcal{A} \text{=} \{ \mathcal{A}_1, \mathcal{A}_2, \ldots, \mathcal{A}_n \}$, where each sub-operation $\mathcal{A}_k : \mathbb{R}^3 \to \mathbb{R}^{d_k}$ independently processes the semantic subspace $\mathcal{P}_k \text{=} \{ p_i \in \mathcal{P} \mid s_i \text{=} k \}$ partitioned from the original point cloud $\mathcal{P} \text{=} \bigsqcup_{k=1}^n \mathcal{P}_k$. Formally, the preprocessing acts as: $\mathcal{A}(\mathcal{P}) \text{=} \left\{ \mathcal{A}_1(\mathcal{P}_1), \mathcal{A}_2(\mathcal{P}_2), \ldots, \mathcal{A}_n(\mathcal{P}_n) \right\}$, where each $\mathcal{A}_k$ maps its assigned semantic subset $\mathcal{P}_k$ to a structured representation $S_k \text{=} \mathcal{A}_k(\mathcal{P}_k) \in \mathbb{R}^{N_k \times d_k}$, with $N_k$ denoting the cardinality of $\mathcal{P}_k$ and $d_k$ the feature dimension.  
(3) The Feature Extractor $\mathcal{F}_k$: $\mathbb{R}^{N_k \times d_k} \to \mathbb{R}^{m_k}$ is used to extract features from $S_k$, producing a feature vector $f_k \text{=} \mathcal{F}_k(S_k) \in \mathbb{R}^{m_k}$. $\mathcal{F}(p)$ denotes the point-level features of point, and $\mathcal{F}(S)$ denotes the feature extraction for each point within the entire semantic space.

\subsection{Theorem}
 Based on the definitions above, the fence theorem formalises the preprocessing action $\mathcal{A}$ as a dual-objective semantic isolator, with the aim of minimising the inter-semantic interference corresponding to each point $\mathcal{P}$ considered by the preprocessing action $\mathcal{A}$, and aligning as much as possible within the semantics, which can be textually stated as (1) Mitigating cross-semantic interference to the greatest extent feasible, and (2) Confining anomaly judgements to aligned semantic spaces wherever viable, thereby establishing intra-semantic comparability. 
 \textbf{All preprocessing approaches are uniformly formalised for achieving this dual goal through a potentially two-stage process, semantic segmentation stage and spatial constraints stage, in which each preprocessing approach is united under the framework of the Fence Theorem, even though it behaves differently at different stages.}
 The existing approaches and our preprocessing approach for Patch3D presented in Secttion~\ref{Patch3D} are shown in Figure~\ref{R4}. This two-stage process can be expressed as follows:
 
 \textbf{Semantic-Division stage.} Given a point cloud $\mathcal{P}$ to be processed, each of its points $p_i$ \text{=} $(x_i, y_i, z_i, s_i)$ has a corresponding true semantic label $s_i$, and $s_i$ is often agnostic before preprocessing. The first step of the preprocessing action $\mathcal{A}$ is to divide the semantic space by partitioning the points that it considers to have the same semantics into the same semantic space $\{\mathcal{P}_k\}$, where $\mathcal{P}$\text{=}$\bigsqcup_{k=1}^n \mathcal{P}_k$. The preprocessing operation will use the corresponding preprocessing operation $A_i$ for each semantic space $P_i$ to get the processed semantic space $S_i$. This process can be formally described as:
 \begin{equation}
     \left\{
     \begin{aligned}
         & \{\mathcal{P}_k\} \text{=}  \mathcal{A}(\mathcal{P}), \mathcal{P} \text{=}\bigsqcup_{k=1}^n \mathcal{P}_k\\
         & S_k \text{=} \mathcal{A}_k(\mathcal{P}_k) \in \mathbb{R}^{N_k \times d_k}, k=1,2,\ldots,n
     \end{aligned}
     \right.
     \label{1}
 \end{equation}

In this process, the semantic space $P_k$ divided by the preprocessing $A$ is processed separately as $S_k$, which guarantees the independence of the subsequent processing. 

\textbf{Spatial-Constraints stage.} The preprocessing action $\mathcal{A}$ tries to ensure that the processed semantic spaces $S_k$ do not influence each other, which means that $S_k$ are as orthogonal as possible to each other. This process can be formally described as:
\begin{equation}
    \left\{
    \begin{aligned}
        & \forall i, j \in \{1, \ldots, n\}, i \neq j: \ 
        tr\left( S_i^\top S_j \right) \text{=} 0 \\
        & \forall i, j \in \{1, \ldots, n\}, i \neq j: \ tr(\mathcal{F}_i(S_i)^\top \mathcal{F}_j(S_j)) \text{=} 0 \\
    \end{aligned}
    \right.
    \label{2}
\end{equation}

where the trace function $tr$ denotes the sum of the diagonal elements of the matrix, and  $tr(\mathcal{F}_i(S_i)^\top \mathcal{F}_j(S_j)) \text{=} 0$ implies that $\mathcal{F}_i(S_i)^\top$ and $\mathcal{F}_j(S_j)$ are orthogonal, preserving independence. 

Through the first and second processes, each preprocessing action $\mathcal{A}$ attempts to divide the point cloud $\mathcal{P}$ into a plurality of mutually non-interfering semantic spaces $S_k$, $k\text{=}\{ 1,\ldots, n \}$. And, these mutually non-interfering semantic spaces $S_k$ still need to be ensured to be mutually non-interfering after being processed in the feature space by their corresponding feature extractors $\mathcal{F}_k$. After these processes, each point $p$ is partitioned into a corresponding semantic space $\mathcal{S}_i$ and is processed by its corresponding preprocessing action $\mathcal{A}_i$.

Finally, during the process of anomaly detection, the preprocessing approach is able to identify the anomaly independently through the use of constrained semantic spaces, which are distinct from one another. The features $f_k$, $k\text{=}1,\ldots,n$, extracted by the feature extractor $\mathcal{F}_k$ are partitioned into different feature spaces by Equations~\ref{1} and~\ref{2}. The feature embedding method and the feature reconstruction method do not operate on the same detection principle. However, a categorisation of these methods can be achieved by comparing the normal structure $p_{nor}$ and the structure to be tested $p_{test}$. This process can be formally described as:
\begin{equation}
    \left\{
    \begin{aligned}
   & AS \text{=} \| \mathcal{F}_{nor}(\mathcal{A}_{nor}(p_{nor})) - \mathcal{F}_{test}(\mathcal{A}_{test}(p_{test})) \|_n \\
   & s_{test} \text{=} s_{nor}
    \end{aligned}
    \right.
    \label{3}
\end{equation}
The subscript test stands for to be tested, $\|( \cdot )\|_n$ represents the $n$-th norm, and $AS$ stands for Abnormal Score. This process indicates that the anomaly detection of the feature to be tested needs to be compared with a normal feature with the same semantics. The fence theorem elucidates the processing flow of any preprocessing approach in anomaly detection and provides a unified guidance process for 3DAD. Existing preprocessing approaches can also be categorised as Fence Theorems, with detailed mathematical reasoning and more extended properties being reported in the \textit{supplementary material}.

\subsection{Evaluation of preprocessing}
The fence theorem provides a unified form of interpretation for preprocessing of 3D anomaly detection. In addition, it is important to qualitatively evaluate the quality of a preprocessing action. In this section, the evaluation of preprocessing actions is divided into two parts: 1) the accuracy $Ac.$ of semantic space segmentation (reasonableness). 2) the number $Nub.$ of semantic space segmentation (fine-grainedness).
\subsubsection{Reasonableness}
\label{d1}
$Ac.$ indicates whether the semantic space $S_i$ into which each point is divided is consistent with its own semantic $s_i$, and if not, the division is considered imprecise. If most of the points are inconsistent, the precision is considered low. Low precision leads to poor anomaly detection performance.

\subsubsection{Fine-Grainedness} The number of semantic spaces divided, denoted $Nub.$, is indicative of the precision of $Ac.$. The greater the division of semantic spaces, the greater the number of points allocated to correct semantics. It is therefore hypothesised that there is a positive correlation between the number of semantic spaces divided and the anomaly detection performance, provided that the division accuracy is guaranteed.

In order to validate the existing approaches through our evaluation methodology, we specifically analysed the ideal situation, the real situation and the existing approaches. It is based on the idea that different semantic spaces should be separate, but current methods often don't do this because they don't have the right features.Two important cases are discussed in the supplementary material: \textbf{Satisfying Constraints} and \textbf{Violating Constraints}. In addition, the results of the analyses for each approach that are available are reported in (textit{supplementary material}).

\section{Approach}
\begin{figure*}[t] 
    \centering 
    \includegraphics[width=1\linewidth]{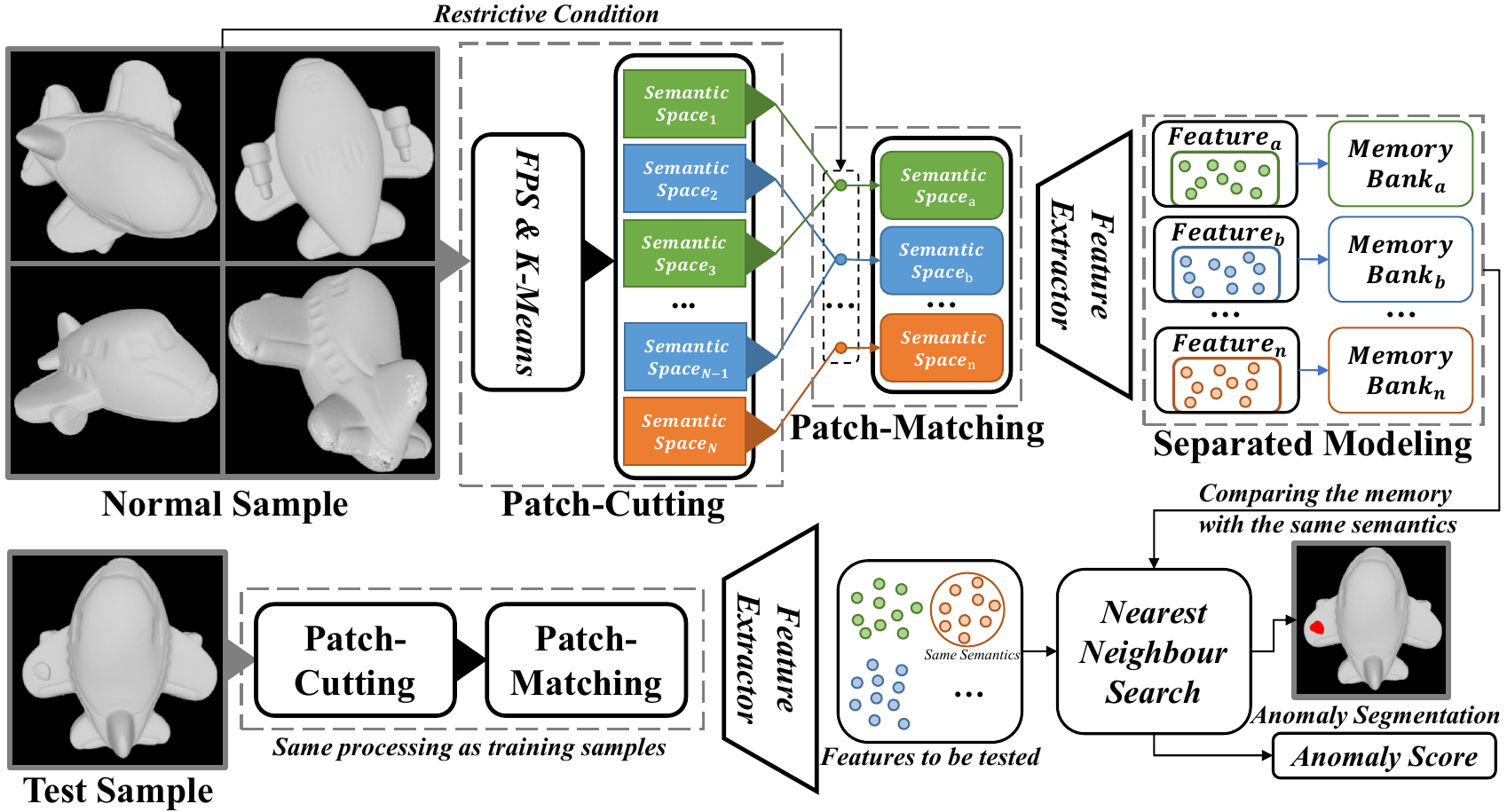} 
    \caption{\textbf{Pipeline of Patch3D.}}
    \label{pipeline}
\end{figure*}
The pipeline of Patch3D consists of three main parts: Patch-Cutting, Patch-Matching and Separation Modeling, as shown in Figure~\ref{pipeline}.
Based on the proposed \textbf{Fence Theorem}, we analyse existing classical preprocessing approaches in this section and propose \textbf{Patch3D}, an approach with full orthogonality between semantic spaces, to verify the correctness of the \textbf{Fence Theorem} in reverse by means of qualitative, quantitative and mathematical analysis.

\subsection{Analysis of existing approaches}
\label{A1}
Existing approaches to preprocessing vary depending on the model and are summarised below: normalisation, normalisation, registration and creation of pseudo exceptions. Existing pre-processing approaches for anomaly detection can be formalised as bimanual semantic isolators and the processing conforms to the two-stage process proposed by the Fence Theorem, a full mathematical representation and an explanation of the implementation will be reported in~\textit{supplementary material}.


\subsection{Patch3D}
\label{Patch3D}
\subsubsection{Motivation}
To verify the Fence Theorem, the goal of our model is to partition the point cloud $\mathcal{P}$ \text{=} $\{p_i\}_{i\text{=}1}^n$
 , where each point $p_i = (x_i, y_i, z_i) \in \mathbb{R}^3 \times \mathcal{L}$,  and $\mathcal{L} \text{=} \{1, 2, \ldots, n\}$, into multiple semantic spaces $\mathcal{S}$\text{=}$\{\mathcal{S}_i\}_{i\text{=}1}^k$ and ensure that the spaces $S_i$ are orthogonal to each other. Our feature extractor utilises the FPFH~\cite{fpfh} features to perform interpretable equality feature extraction for each point, which can be defined as $\mathscr{F}_4: \mathbb{R}^3\rightarrow \mathbb{R}^{33}$, meaning that the point coordinates are mapped to a feature vector of dimension 33. This goal can be formally described as:
\begin{equation}
    \left\{
    \begin{aligned}
    & \forall i, j \in \{1, \ldots, k\}, i \neq j: \ 
    tr\left( \mathcal{S}_i^\top \mathcal{S}_j \right) \text{=}\ 0 \\
    & \forall i, j \in \{1, \ldots, k\}, i \neq j: \ tr(\mathscr{F}_4(\mathcal{S}_i)^\top \mathscr{F}_4(\mathcal{S}_j)) \text{=}0 \\
    \end{aligned}
    \right.
\end{equation}
This process implies that the semantic space delineated by our Patch3D approach and the feature semantic space created by subsequent representations are orthogonal to each other, in line with the motivation of the Fence Theorem.

\subsubsection{Patch-Cutting}
We propose the patch-cutting approach to split a complete point cloud $\mathcal{P}$ into multiple parts to create multiple semantic spaces $\hat{\mathcal{S}}$\text{=}$\{\mathcal{S}_i\}_{i\text{=}1}^k$. This process relies only on the structure of a single point cloud itself for segmentation, without interaction between multiple point clouds. First, we select the farthest point from the centre of gravity of the point cloud as the starting point using Farthest Point Sampling~(FPS)~\cite{fps} to obtain a downsampled set of points $\hat{\mathcal{P}}$ \text{=} $\{\hat{p}_i\}_{i\text{=}1}^k$. Then, we use the K-means algorithm to partition the point cloud $\mathcal{P}$ into multiple unaligned semantic spaces $\mathcal{S}_i$ based on the clustering of the set of points $\hat{\mathcal{P}}$. This Patch-Cutting can be formally described as:
\begin{equation}
    \left\{
    \begin{aligned}
    &\{\hat{p}_i\}_{i\text{=}1}^k \text{=} FPS(\mathcal{P}, start\ point\text{=}p_{str})\\
    & p_{str}\text{=}\arg\max_{p \in \mathcal{P}} \| p\text{-} c_{ctr} \|, c_{ctr} \text{=} \frac{1}{n} \sum^{n}_{i=1} p_i \\
    &\{\mathcal{S}_i\}_{i\text{=}1}^k \text{=} K\text{-}Means(\{\hat{p}_i\}_{i\text{=}1}^k)
    \end{aligned}
    \right.
\end{equation}
In addition, to ensure that the division is sufficiently homogeneous, we impose an additional constraint on K-Means: the number of points in each semantic space must not vary too much. Assume that the number of points corresponding to each of the semantic spaces $\{ \hat{\mathcal{S}}_1,\hat{\mathcal{S}}_2,\ldots,\hat{\mathcal{S}}_N \}$ is $\{ \alpha_1, \alpha_2, \ldots, \alpha_N \}$. This additional constraint can be formally described as:
\begin{equation}
    \forall i,j \in \{1,2,\ldots,N\}, \alpha_i \le \delta \cdot \alpha_j
\end{equation}
where $\delta$ represents an equilibrium parameter, i.e. a tolerance value for the maximum multiple of the difference between points. Through the preprocess of Patch-Cutting, we can obtain preliminary semantic labels for each point. This process can be formally expressed as:
\begin{equation}
    \forall p_i,i\in\{1, 2, \ldots, n\},\exists \mathcal{S}_j,j\in\{1, 2, \ldots, k\},\ p_i \in \mathcal{S}_j.
    \label{9}
\end{equation}
During the Patch-Cutting, we processed a single point cloud $\mathcal{P}$. Similarly, we processed all the training and test point clouds $\{ \mathcal{P}_1,\mathcal{P}_2,\ldots,\mathcal{P}_N \}$, each of which is divided into a corresponding set of semantic spaces $\{ \hat{\mathcal{S}}_1,\hat{\mathcal{S}}_2,\ldots,\hat{\mathcal{S}}_N \}$.

\subsubsection{Patch-Matching}
In the Patch-Cutting phase, the semantic spaces delineated for each point cloud are independent of each other. This process ensures that the semantic spaces are orthogonal to each other within the point cloud. However, it lacks the merging of semantic spaces between the point clouds. Consequently, each point cloud is delineated into multiple similar semantic space, but the meanings of the semantic spaces are not aligned. To merge the independent sets of semantic spaces $\{ \hat{\mathcal{S}}_1,\hat{\mathcal{S}}_2,\ldots,\hat{\mathcal{S}}_N \}$ of multiple point clouds, Patch-Matching is proposed to compute the distance from the centre of mass of the points contained in each semantic space of a single point cloud to the centre of mass of its point cloud and obtain a multiple descending order. 
\begin{equation}
    \left\{
    \begin{aligned}
    & c_i\text{=}\frac{1}{\alpha_i} \sum^{\alpha_i}_{j=1} p_j, i\in \{ 1,2,\ldots,k \}\\
    & d_i\text{=}||c_i \text{-} c_{ctr}||,i\in \{1,2,\ldots,k\}\\
    & Order(\{\mathcal{S}_i\}_{i\text{=}1}^k)\text{=}Des\text{-}Order(\{d_i\}_{i\text{=}1}^k)
    \end{aligned}
    \right.
\end{equation}
where $Des\text{-}Order$ denotes descending order, the semantic space is ordered according to the distance of the centroid of the points it contains from the total centroid of the point cloud, with the closer the distance, the higher the order number. According to each element in this logical semantic space set $\{ \hat{\mathcal{S}}_1,\hat{\mathcal{S}}_2,\ldots,\hat{\mathcal{S}}_N \}$, its internal ordering is re-specified. At this point, semantic spaces with the same ordinal number in different elements are merged and treated as having the same semantics. For example, the semantic space in which each element in the semantic set is ranked first, i.e. closest to the centre of mass, is merged and denoted as $\overline{\mathcal{S}}_1$. Following this logic, all semantic spaces are merged and the merged denoted as $\{\overline{\mathcal{S}}_i\}_{i\text{=}1}^N$\text{=}$\{ \overline{\mathcal{S}}_1,\overline{\mathcal{S}}_2,\ldots,\overline{\mathcal{S}}_N \}$. Assume that the number of points in each of the merged semantic spaces is $\{\overline{\beta}_i\}_{i\text{=}1}^N$\text{=}$\{ \beta_1, \beta_2, \ldots, \beta_N \}$.

Following the implementation of Patch-Cutting and Patch-Matching, the point cloud is then partitioned into multiple semantic spaces $\{\overline{\mathcal{S}}_i\}_{i\text{=}1}^N$. The locations of similar point clouds are then partitioned into the same semantic space. Furthermore, the semantic spaces $\{\overline{\mathcal{S}}_i\}_{i\text{=}1}^N$ are orthogonal to each other, as they are explicitly labelled differently, in line with the motivation of the \textbf{Fence Theorem}.

\subsubsection{Separated Modeling}
In order to maintain the mutual orthogonality between the semantic spaces to the mutual orthogonality of the feature spaces, we embed each point $\{p_i\}_{i\text{=}1}^n$ in the training point cloud $\mathcal{P}$ into the feature memory $\mathcal{M}$ corresponding to the semantics to which it belongs according to Equation~\ref{9}. The process of embedding the features of each point into the memory bank can be formalised:
\begin{equation}
    \mathcal{M}_i \text{=} \{\mathscr{F}_4(p_k) \mid p_k \in \overline{\mathcal{S}_i}, k\in \{1, 2, \dots, \beta_i\}, i \in \{ 1, 2, \ldots, N\} \}
\end{equation}
Through this process, we embedded all the points of the training point cloud into the memory bank. Specifically, the features of each point contained in the semantic space $\{\overline{\mathcal{S}}_i\}_{i\text{=}1}^N$ are embedded into its corresponding semantic memory $\{{\mathcal{M}}_i\}_{i\text{=}1}^N$\text{=}$\{ \mathcal{M}_1, \mathcal{M}_2, \ldots, \mathcal{M}_N \}$, separately. In this process, the memory bank $\{{\mathcal{M}}_i\}_{i\text{=}1}^N$ contains the corresponding $\{\overline{\beta}_i\}_{i\text{=}1}^N$\text{=}$\{ \beta_1, \beta_2, \ldots, \beta_N \}$ feature vectors, which means that the normal features are modelled separately.  

In the testing phase, the points $p_{test}$ to be evaluated merely require comparison with their semantic counterparts in the memory. Following the guidance of Equation 1,the anomaly detection process for the feature to be tested can be formalised as follows:
\begin{equation}
    \mathcal{AS} \text{=} \|\mathscr{F}_4(p_{test}) - \mathscr{F}_4(p_{neb})\|_2
\end{equation}
where $\mathcal{AS}$ represents the anomaly score, and $p_{neb}$ represents the nearest neighbour to the feature to be tested in the memory bank. $p_{test}$ and $p_{neb}$ \textbf{must} belong to the same semantic space. This Separated Modeling and making the points to be detected anomalous in their semantic space is in accordance with the Fence Theorem, which is centred on making feature spaces with different semantics orthogonal and independent of each other.

\section{Experiments}
\subsection{Experimental Settings}
\textbf{Datasets.} Our evaluation is performed on two anomaly detection datasets Real3D-AD and Anomaly-ShapeNet. Real3D-AD is an anomaly detection dataset from a high-precision scanning device with twelve classes, each having four normal samples for training and over a hundred samples for testing. Anomaly-ShapeNet is a synthetic dataset from ShapeNet datasets, with 40 classes and a total of over 1600 point cloud samples for anomaly detection evaluation.

\textbf{Evaluation Metrics.} For the anomaly detection task, we use Area Under the Receiver Operating Characteristic Curve~(AUROC) and Area Under the Precision versus Recall Curve~(AUPR) for evaluation. In particular, O-AUROC~($\uparrow$) and O-AUPR~($\uparrow$) are used to evaluate object-level anomaly detection capability, and P-AUROC~($\uparrow$) and P-AUPRO~($\uparrow$) are used to evaluate point-level anomaly detection capability. Higher values of these four metrics indicate better anomaly detection capability.

\textbf{Baseline.} In this study, the objective is to evaluate the reasonableness of previous preprocessing algorithms under the Fence Theorem. For the registration approach, we have selected PatchCore (FPFH + Raw)~\cite{Patchcore}, Reg3D-AD~\cite{Real3D-AD} and ISMP~\cite{ISMP}; for the pseudo-anomaly generation approach, we have opted for the Patch-Gen~\cite{Real3D-AD} approach with Norm-AS~\cite{PO3AD} to replace the preprocessing approach in PO3AD; for the standardisation and normalisation, we have utilised the Raw and FPFH features to observe the features offset. It is noteworthy that all approaches have been derived from open-source code or reproduced results.

\subsection{Evaluation of Existing Approaches}
\subsubsection{Registration Approach}
\begin{figure}[t] 
    \centering 
    \includegraphics[width=1\linewidth]{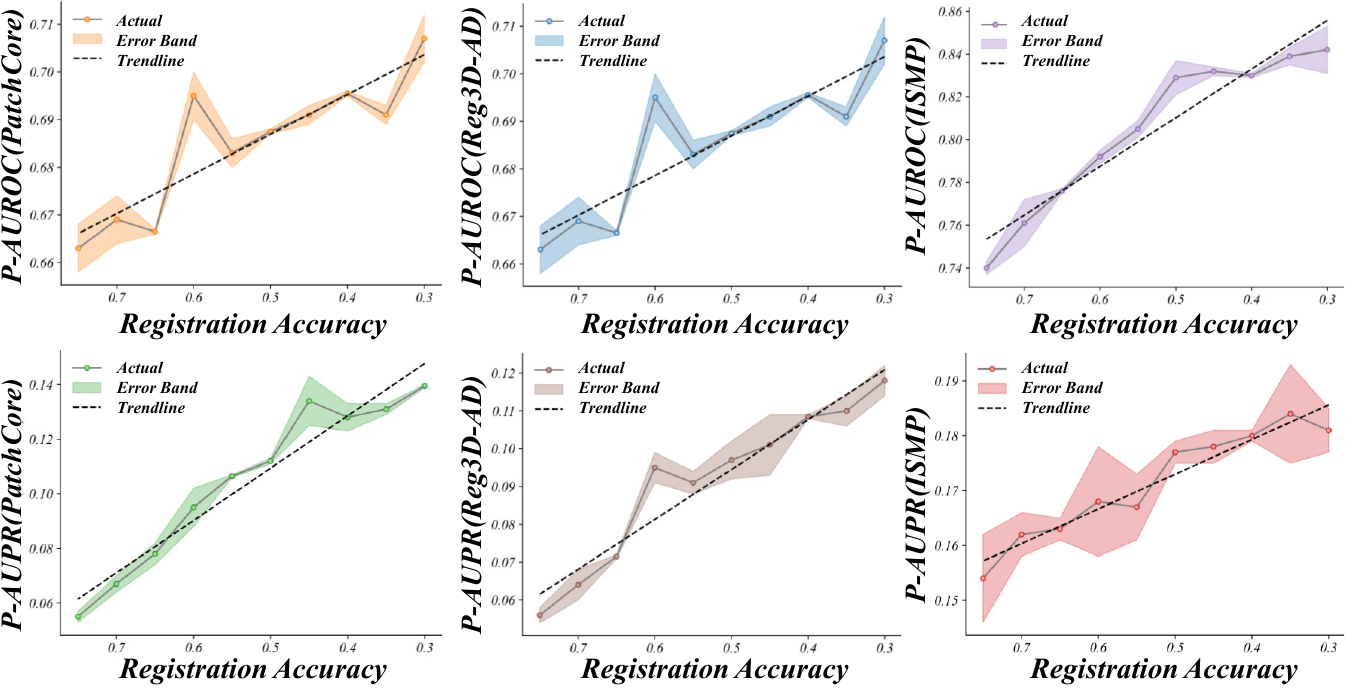} 
    \caption{\textbf{Experimental results of the registration approach.} There is a significant positive correlation between the improved point-level detection performance and the registration accuracy. This is attributed to the increased orthogonality of the individual semantic spaces in the presence of increased registration accuracy.}
    \label{R1}
\end{figure}
Experiments on the alignment approaches are performed by modifying the \textit{voxel size} in the RANSAC algorithm~\cite{fpfh}. It has been demonstrated that an excessively high \textit{voxel size} can result in a loss of point cloud accuracy, a decrease in the registration accuracy, and a non-orthogonal division of the semantic space. This, as mentioned in Section~\ref{A1}, can have a negative impact on anomaly detection. The results of the experiments are displayed in Figure~\ref{R1}, where all the approaches for anomaly detection using registration show a rise in detection ability as the alignment accuracy increases, manifested as a rise in point-level anomaly detection ability.  This is attributable to the fact that as the registration accuracy improves, the point-level features of the point cloud to be tested can be more readily compared with the point-level features in the training point cloud that possess the same semantics, thereby facilitating the detection of anomalies. The complete experimental data and model efficiencies are presented in the 
\textit{supplementary material}.

\subsubsection{Pseudo-Anomaly Generation Approach}
\begin{table}[htbp]
\resizebox{\linewidth}{!}{
   \begin{tabular}{cc|cc|c|cc}
    \hline
    \multicolumn{4}{c|}{\textbf{Norm-AS}} & \multicolumn{3}{c}{\textbf{Patch-Gen}} \\
    \hline
    \multicolumn{2}{c|}{\textbf{Scaling Factor}} & \multicolumn{2}{c|}{\textbf{Results}} & \textbf{Scaling Factor} & \multicolumn{2}{c}{\textbf{Results}} \\
    \cline{1-4} \cline{5-7}
    Range & Mean & P-AUROC & P-AUPR & Value & P-AUROC & P-AUPR \\
    \hline
    0.06-0.12 & 0.09 & 87.43\% & 44.87\% & 0.1 & 63.26\% & 3.38\% \\
    0.05-0.11 & 0.08 & 87.40\% & 44.99\% & 0.2 & 62.75\% & 3.27\% \\
    0.04-0.10 & 0.07 & 85.57\% & 41.27\% & 0.3 & 61.64\% & 3.24\% \\
    0.03-0.09 & 0.06 & 83.81\% & 39.54\% & 0.4 & 59.77\% & 2.49\% \\
    0.02-0.08 & 0.05 & 81.43\% & 37.28\% & 0.5 & 57.42\% & 2.13\% \\
    0.01-0.07 & 0.04 & 74.41\% & 31.01\% & 0.6 & 54.14\% & 1.21\% \\
    0.00-0.06 & 0.03 & 71.27\% & 29.87\% & 0.7 & 52.49\% & 1.30\% \\
    \hline
\end{tabular}%
}
\caption{\textbf{Experimental results of pseudo-Anomaly generation approach}
It can be observed that the closer the creation of pseudo anomalies is to the real one (i.e. the closer the undulations are to the real anomalies), the better the anomaly detection effect of the model is. All experiments were done on two RTX 2080Ti.
\label{b1}}
\end{table}
It is hypothesised that the anomalies created by the proposed Norm-AS and Patch-Gen approaches are similar to real anomalies, and that these anomalies are transformed in a more spurious direction by adjusting the scaling factor. The results demonstrate that the generated anomalies become increasingly inconspicuous as the scaling factor varies, and the point-level anomaly detection results deteriorate, as demonstrated in Table~\ref{b1}. For instance, the Norm-AS approach reduces the P-AUROC and P-AUPR by 0.1616 and 0.1507, respectively, as the mean value of the factor changes from 0.9 to 0.3. This finding aligns with our Fence Theorem, which posits that when anomalies are constrained by semantic spatial orthogonality, the anomaly detection becomes less effective.
\subsubsection{Standardisation and Normalisation Approach}

\begin{table*}[htbp]
\resizebox{\textwidth}{!}{
    \begin{tabular}{c|cccccccccccc|cc}
    \hline
    \multicolumn{15}{c}{\textbf{Raw}} \\
    \hline
    \textbf{Method} & \textbf{airplane} & \textbf{car} & \textbf{candybar} & \textbf{chicken} & \textbf{diamond} & \textbf{duck} & \textbf{fish} & \textbf{gemstone} & \textbf{seahorse} & \textbf{shell} & \textbf{starfish} & \textbf{toffees} & \multicolumn{2}{c}{\textbf{Mean}} \\
    \hline
    \textbf{Origin} & 1.4074/2.1116 & 1.3140/0.3667 & 1.7321/0.2229 & 4.5481/0.6384 & 3.9614/0.6171 & 0.9865/1.4332 & 0.5820/0.3705 & 2.8208/0.4166 & 0.3983/0.1779 & 2.8347/0.4358 & 4.0910/1.3783 & 0.8439/0.2818 & \multicolumn{2}{c}{2.1267/0.7042} \\
    \textbf{Normalisation} & 0.0493/0.0218 & 0.0422/0.0057 & 0.0659/0.0099 & 0.1239/0.0326 & 0.0521/0.0124 & 0.0763/0.0484 & 0.0306/0.0135 & 0.0589/0.0112 & 0.0195/0.0031 & 0.0527/0.0124 & 0.0698/0.0064 & 0.0501/0.0126 & \multicolumn{2}{c}{0.0576/0.0158} \\
    \hline
    \multicolumn{15}{c}{\textbf{FPFH}} \\
    \hline
    \textbf{Method} & \textbf{airplane} & \textbf{car} & \textbf{candybar} & \textbf{chicken} & \textbf{diamond} & \textbf{duck} & \textbf{fish} & \textbf{gemstone} & \textbf{seahorse} & \textbf{shell} & \textbf{starfish} & \textbf{toffees} & \multicolumn{2}{c}{\textbf{Mean}} \\
    \hline
    \textbf{Origin} & 1.4913/1.3206 & 9.7578/7.7984 & 1.5643/1.0561 & 2.7259/3.2247 & 3.3104/5.8938 & 0.5196/1.1942 & 1.9146/2.2580 & 1.8569/0.9300 & 0.8173/0.9126 & 1.8861/1.1770 & 2.3910/1.1710 & 0.7220/1.5204 & \multicolumn{2}{c}{2.4134/2.3714} \\
    \textbf{Normalisation} & 0.0109/0.0165 & 0.0799/0.0773 & 0.0147/0.0151 & 0.0259/0.0474 & 0.0280/0.0561 & 0.0025/0.0081 & 0.0180/0.0307 & 0.0172/0.0168 & 0.0105/0.0117 & 0.0132/0.0094 & 0.0169/0.0109 & 0.0083/0.0169 & \multicolumn{2}{c}{0.246/0.0264} \\
    \textbf{Standardization } & 0.0002/0.0015 & 0.0048/0.0016 & 0.0001/0.0004 & 0.0001/0.0012 & 0.0484/0.0160 & 0.0286/0.0074 & 0.0001/0.0007 & 0.0001/0.0004 & 0.0001/0.0004 & 0.0001/0.0008 & 0.0001/0.0006 & 0.0002/0.0018 & \multicolumn{2}{c}{0.0069/0.0027} \\
    \hline
    \end{tabular}%
}
\caption{\textbf{Please find below a comparison between the Standardisation,  Normalisation and the origin point cloud.}
\textit{a/b} represents the mean difference and variance difference between the training and test sets, respectively. The specific value of variance is the mean value of each dimension of the feature. All experiments were done on two RTX 2080Ti.
\label{nomor}}
\end{table*}

Standardisation and normalisation are standard operations before extracting features from a point cloud, and normalisation is usually used to make point clouds of the same scale, projected into a comparable primary semantic space. As shown in Table~\ref{nomor}, the point cloud and its features without any preprocessing are very discrete, and the difference between the training and test sets is large, with a difference of 2.1267 in the mean coordinates and 0.7042 in the variance, in addition to a difference of 2.4134 in the mean and 2.3714 in the variance of the FPFH features. the feature distributions are tightened up significantly and the difference between the training and test sets becomes smaller after simple normalisation and normalisation, e.g., the normalised The difference between the coordinate means of the training set and the test set after normalisation is only 0.0576, and the variance is also reduced to 0.0158. this means that normalisation allows the point cloud to be mapped into a more comparable space, restricting the anomaly detection to the global point cloud in line with Fence Theorem.

\subsection{Evaluation of Patch3D}
\subsubsection{Main Results}
\label{mainresults}
\begin{figure}[t] 
    \centering 
    \includegraphics[width=1\linewidth]{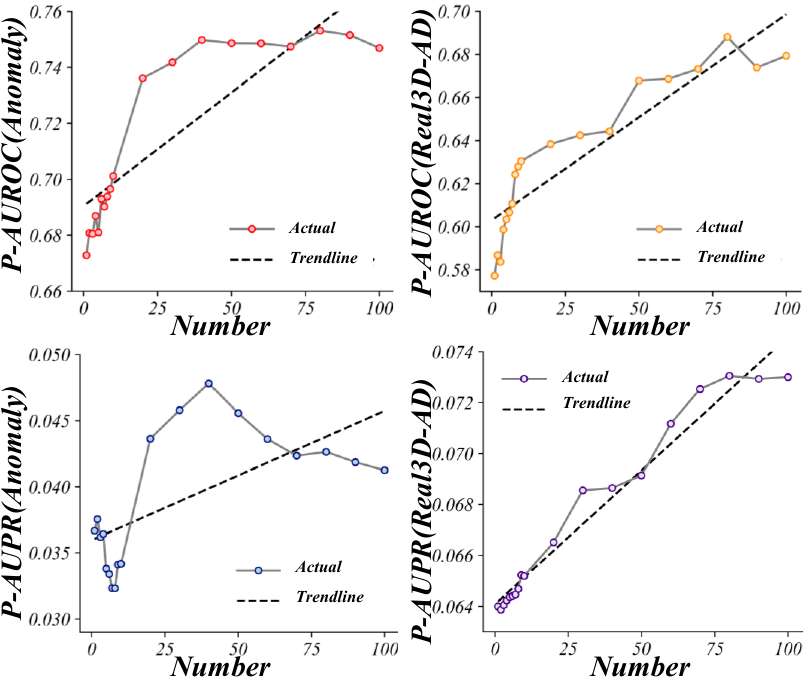} 
    \caption{\textbf{Experimental results of the Patch3D.} There is a significant positive correlation between the improvement in point-level detection performance and the number of semantic spaces. This is consistent with the interpretation of the Fence Theorem.\label{R2}}
\end{figure}
The correspondence between the number of semantic spaces divided and the point-level anomaly detection performance can be reversed to verify the nature of the Fence Theorem in Section~\ref{d1}. The validation on Real3D-AD and Anomaly-ShapeNet are shown in Figures~\ref{R2}. The performance of point-level anomaly detection, measured by P-AUROC and P-AUPR, exhibited a clear positive correlation trend with the semantic space division across both datasets. Notably, on the Real3D-AD dataset, P-AUROC and P-AUPR increased by 0.1108 and 0.0091, respectively, with the increase in semantic space division. The P-AUROC and P-AUPR of Anomaly-ShapeNet increased by 0.0803 and 0.0111, respectively, and the P-AUPR exhibited an initial increase followed by a subsequent decrease. The discussion of this trend is provided in the Limitation section.The overall growth trend is consistent with the conclusion of our Fence Theorem. Complete experimental data are presented in the \textit{Supplementary Material}.

\subsubsection{Efficiency analysis}
\begin{figure}[t]
    \centering
    \includegraphics[width=0.95\linewidth]{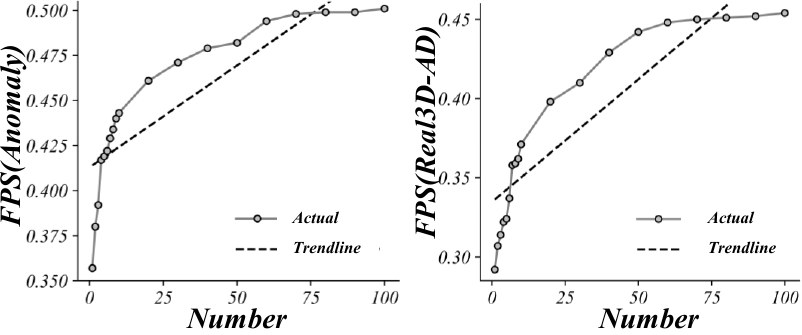} 
    \caption{Quantitative relationship between model rate and semantic space division.}
    \label{R3}
\end{figure}

\begin{figure}[t]
    \centering
    \resizebox{0.75\linewidth}{!}{ 
        \begin{tabular}{c|cc}
        \hline
        Method & Time Complexity & Orthogonality \\
        \hline
        Reg3D-AD & \textit{O(n)} & \XSolidBrush  \\
        Group3AD & \textit{O(n)} & \XSolidBrush \\
        M3DM  & \textit{O(n)} &  \XSolidBrush \\
        BTF   & \textit{O(n)} &  \XSolidBrush \\
        ISMP  & \textit{O(n)} &  \XSolidBrush \\
        Patch3D(K) & \textit{O(n/K)} & \CheckmarkBold \\
        \hline
        \end{tabular}%
    }
    \caption{Comparison of time complexity and feature orthogonality.}
    \label{b2}
\end{figure}

The time complexity of Patch3D is shown to be inferior to that of other feature-embedding approaches that are already available, as demonstrated quantitatively in Figure~\ref{R3}. As the semantic space increases, the FPS of anomaly detection rises, which is due to the fact that the features to be tested need to be compared with fewer features and the time efficiency increases. Specifically, as shown in Figure~\ref{R4}, existing approaches do not differentiate the semantic space, and the features to be tested need to compare all the features in the memory bank, whereas our approach only needs to compare the feature memories with the same semantics. Removing the time difference due to the feature extractor, we qualitatively analysed in Table~\ref{b2}, assuming that $K$ semantic spaces are partitioned and have the same number of features within each semantic space. The number of features it needs to compare changes from $n$ to $n/K$, and the time complexity decreases from $O(n)$ to $O(n/K)$. Complete experimental data are reported in the \textit{Supplementary Material}.

\subsubsection{Limitations}
The Patch3D model is employed to demonstrate the impact of preprocessing limitations on anomaly detection performance. As discussed in section~\ref{mainresults}, the P-AUPR demonstrates an initial upward trend, followed by a subsequent downward trend. The sample-level test performance, presented in the \textit{supplementary material}, exhibits erratic behaviour. This phenomenon can be attributed to the following factors and it is \textbf{universal} in 3D anomaly detection, and we will further explain the reasons for these limitations in the supplementary material.

\noindent1) The presence of noise in the point cloud, in conjunction with the constraints imposed by the preprocessing approach, results in the semantic space into which the points are divided deviating from their actual semantics.

\noindent2) The noise in the point cloud itself lacks sufficient semantic representation, and the influence of the noise gradually increases after the semantic space is divided into more.

\noindent3) Anomaly detection is determined only by relying on the maximum value score at the point level, which greatly increases the likelihood of false detections caused by the noise.

\section{Conclusion}
In this paper, we proposal the Fence Theorem, which formalises all preprocessing as a bi-objective semantic isolator for problems related to interpretability: mitigating cross-semantic interference and restricting anomalous judgments to the aligned semantic space. We generalise existing preprocessing approaches through qualitative analysis, quantitative verification and mathematical proofs. To validate the Fence Theorem, we develop Patch3D, which includes Patch-Cutting and Patch-Matching modules to decouple semantic spaces and separated model normal features independently in each space. Experiments conducted in Anomaly-ShapeNet and Real3D-AD show that fine semantic alignment in preprocessing improves the accuracy of point-level anomaly detection, and all experimental results point to the correctness of the Fence Theorem.

{
    \small
    \bibliographystyle{ieeenat_fullname}
    \bibliography{main}
}
\clearpage
\section{Supplementary Material}
\begin{figure*}[t] 
    \centering 
    \includegraphics[width=0.8\textwidth]{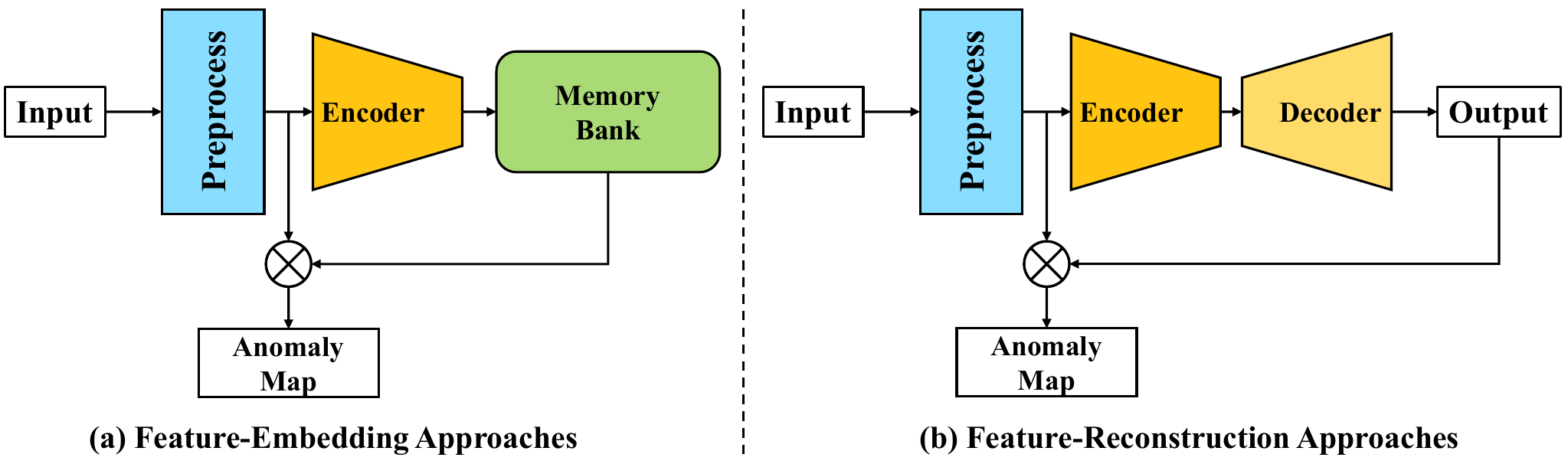} 
    \caption{Comparison of architectures. }
    \label{Comparison}
\end{figure*}
This supplementary material provides the series of elements mentioned in the text: 1) a mathematical formalisation of the process of generalising existing preprocessing approaches under the Fence Theorem. 2) more additions to the Fence Theorem. 3) more additions and analyses of the experimental data. This supplementary material helps the reader to better understand the fence theorem, and we have divided the supplementary material into three points, indicated below by a table of contents with link jumps:

\begin{itemize}
    \item \hyperref[sec1]{A Mathematical Formalisation of an Existing Preprocessing Approach}.
    
    \item \hyperref[sec2]{More Additions to the Fence Theorem}.
    
    \item \hyperref[sec3]{More Results and Analyses of Experimental Data}.
\end{itemize}

Note that the serial numbers of the formulas mentioned here are consistent with the text, and we no longer number the formulas starting from 1. In addition, we provide here all the symbols used in the text and in the supplementary material, together with their explanations, as shown in the table below:

\subsection{A Mathematical Formalisation of an Existing Preprocessing Approach}
\label{sec1}
We report in this section how each preprocessing approach is specifically formalised as a semantic structure isolator as described in the Fence Theorem. Specifically, we mathematically describe how the 1) Creating Pseudo-Anomalies, 2) Registration Approach, 3) Normalization and Standardization, 4) Patch3D are described by the Fence Theorem. Furthermore, the feature-embedding and feature-reconstruction based approach is shown in Figure~\ref{Comparison} to help better understanding. (1) Mitigating cross-semantic interference to the greatest extent feasible and (2) Confining anomaly judgements to aligned semantic spaces wherever viable, thereby establishing intra-semantic comparability are denoted as \textbf{Goal1} and \textbf{Goal2}, respectively. Semantic-Division stage and Spatial-Constraints stage are denoted as \textbf{Stage1} and \textbf{Stage2} respectively

\subsubsection{Creating Pseudo-Anomalies}
Creating pseudo-anomalies refers to the use of mathematical or deep learning approaches to transform otherwise normal structures into anomalous ones, so that the model recognises the compositional differences between the anomalous and normal structures. Creation of Pseudo-Anomalies is often used prior to feature reconstruction, explicitly creating two semantic spaces, normal and anomalous, such that the model imposes different preprocessing actions on the two semantic spaces.

First we describe the concrete representation of the dual\text{-}objective of the pseudo-anomaly generation approach in Fence Theorem. Suppose that the pseudo-anomaly setting is created as $\mathcal{A}\text{=}\{\mathcal{A}_{Ano},\mathcal{A}_{Nor}\}$ and the point cloud being processed is $\mathcal{P}$\text{=}$\{p_1,p_2,\ldots,p_n\}$. Selecting a subset as an anomaly generating point $\mathcal{P}_{Ano}$=$\{p_1,p_2,\ldots,p_i\}$, hence the normal point is expressed as $\mathcal{P}_{Nor}$ \text{=} $\mathcal{P}$ \text{-} $ \mathcal{P}_{Ano}$\text{=}$\{p_1,p_2,\ldots,p_j\}$, where $p_i$=$(x_i, y_i, z_i, s_i)$. This means that the original point cloud has $n$ points, the abnormal part has $i$ points, the normal part has $j$ points, and the abnormal and normal parts are processed by $\mathcal{A}_{Ano}$ and $\mathcal{A}_{Nor}$, respectively. Semantic-Division stage and Spatial-Constraints stage are denoted as $Stage1$ and $Stage2$ respectively

\noindent\textbf{Goal1} This goal aims to make the model learn the ability to minimise the mutual interference between the structures of the model's abnormal semantic space $\mathcal{S}_{Ano}$ and normal semantic space $\mathcal{S}_{Nor}$ during testing. Specifically, the model is trained by first selecting $\mathcal{P}_{Ano}$ and creating the anomalous semantic space via $ \mathcal{A}_{Ano}(\mathcal{P}_{Ano}) = \mathcal{S}_{Ano}$ so that the model learns the ability to reconstruct points within the semantic space $\mathcal{S}_{Ano}$ as normal points with the same semantics $s_i$ as it. Note that by reconstructing as a normal point with the same semantics we mean the point $p_i$=$(x_i, y_i, z_i, s_i)$, the $s_i$ it contains, and not the semantic space $\mathcal{S}_{Nor}$ or $\mathcal{S}_{Ano}$ it belongs to

\noindent\textbf{Goal2} This goal requires that in the test set, anomalous structures are reduced to normal structures or poorly reconstructed parts that need to be compared with their normal structures with the same semantics in order to accurately judge the anomalies. This is due to the fact that the model learns to reconstruct structures within the anomalous semantic space as normal structures with the same semantic $s_i$, which limits the anomaly judgement to within points with the same semantics.

\noindent\textbf{Stage1}
In this stage, the point cloud is segmented into multiple semantic spaces. Specifically, the point cloud is divided into two semantic spaces by applying the preprocessing operation $\mathcal{A}\text{=}\{\mathcal{A}_{Ano},\mathcal{A}_{Nor}\}$ . We can obtain $\{\mathcal{P}_{Ano}, \mathcal{P}_{Nor}\}$\text{=}$\mathcal{A}(\mathcal{P})$ and the resulting semantic spaces, denoted as $\mathcal{S}_{Nor}$ and $\mathcal{S}_{Ano}$, are obtained through $\mathcal{S}_{Nor} = \mathcal{A}(\mathcal{P}_{Nor})$ and $\mathcal{S}_{Ano} = \mathcal{A}(\mathcal{P}_{Ano})$, respectively. $\mathcal{S}_{Ano}$ corresponds to the anomalous semantic space, where points are distorted to deviate from the normal pattern. $\mathcal{S}_{Nor}$ corresponds to the normal semantic space, where points remain unchanged. Through this operation, the point cloud is divided into multiple semantic parts, and each part is processed separately.

\noindent\textbf{Stage2}
The Spatial-Constraints that create the pseudo-anomaly approach are implemented via a loss function. Specifically, since the pseudo-anomaly space $\mathcal{S}_{Ano}$ is specially processed during training so that it deviates from the normal pattern, while the normal space $\mathcal{S}_{Nor}$ is not processed, this leads to the agent task of model training focussing on returning the anomalies back to normal, which makes it necessary for the model to have the ability to maintain the distribution of points in the normal semantics, as well as the ability to restore points in the anomalous semantics. This process is implemented by the mapping function $\mathscr{F}_1: F_{ori} \rightarrow F_{rec}$ mentioned in the \textit{Introduction} section. Through this process, the model handles normal and abnormal semantic structures without interfering with each other, and since the model successfully eliminates the anomalies, the anomaly judgements are restricted to the same semantic space.

\subsubsection{Registration Approach}
The registration approach refers to the use of mathematical or deep learning methods to adjust the poses of similar point clouds to the same orientation so that the model learns to ignore structural information beyond the pose. Registration preprocesses data before feature embedding by aligning poses between semantically identical structures, transforming their coordinates into proximity representations. While struggling with cross-semantic feature interference, it validates our theorem.

Suppose that the point cloud to be processed for the registration approach is $\mathcal{P}$\text{=}$\{p_1,p_2,\ldots,p_n\}$ and its corresponding preprocessing action is $\mathcal{A}$\text{=}$\{\mathcal{A}_1,\mathcal{A}_2,\ldots,\mathcal{A}_n\}$. This means that each point is processed by a different preprocessing action. Assume that the post-processing point cloud $\hat{\mathcal{P}}$\text{=}$\{\hat{p}_1,\hat{p}_2,\ldots,\hat{p}_n\}$, which means that a total of $n$ points are processed and the number of points before and after processing is constant.

\noindent \textbf{Goal1} The first objective of the registration method is to minimize mutual interference between different semantic spaces. Specifically, the model needs to represent structures with the same semantics in the training and test sets with identical features, while features from different semantic spaces should be orthogonal (distinct). Through registration, the feature extractor ensures that similar structures with the same semantics have identical representations, eliminating the impact of rotation on representations and preventing interference between semantic spaces. Ideally, $\hat{p}_i$, representing the $i$-th point in the registered point cloud $\hat{\mathcal{P}}$, follows a consistent distribution for points with the same semantics, while distributions between different semantics remain orthogonal, thus eliminating interference between different semantics.

\noindent \textbf{Goal2} Building on Goal 1, Goal 2 manifests as confining anomaly judgments to the space with the same semantics. That is, during anomaly judgment, the model needs to compare with the distribution of the same semantics as much as possible, reducing the influence from other semantic distributions. In non-ideal cases, such as ISMP, the semantic spaces are not orthogonal, and anomaly judgments will be affected by other distributions, resulting in inaccurate anomaly judgments. When the semantic spaces are orthogonal, each feature distribution is completely isolated. In this case, anomaly judgments are confined to the scope with the same semantics, achieving precise anomaly detection.

\noindent \textbf{Stage1}
In this stage, the point cloud is divided into $m$ semantic spaces. Specifically, the preprocessing operation $\mathcal{A}$ is applied to segment the point cloud $\mathcal{P}$ into multiple point sets $\{\mathcal{P}_1, \mathcal{P}_2, \ldots, \mathcal{P}_m\}$ \text{=} $\mathcal{A}(\mathcal{P})$. The resulting semantic spaces are denoted as $\mathcal{S}_1, \mathcal{S}_2, \ldots, \mathcal{S}_m$, where each semantic space contains points with similar structures and semantic information. Here, $\mathcal{S}_i$ \text{=} $\mathcal{A}_i(\mathcal{P}_j)$ for $i$ \text{=} $1, 2, \ldots, m$, and each point $p_i$ is assigned to a specific semantic space and processed by the corresponding preprocessing operation $\mathcal{A}_i$. The registration process aligns points within the same semantic space, reducing interference between different spaces and enabling the model to focus on relevant structural information. This stage produces registered point clouds $\{\hat{\mathcal{P}}_1, \hat{\mathcal{P}}_2, \ldots, \hat{\mathcal{P}}_m\}$, laying the foundation for the subsequent stage where features from different semantic spaces will be made as orthogonal as possible.

\noindent \textbf{Stage2}
The registered point clouds are processed with the goal of making the features from different semantic spaces as orthogonal as possible. Specifically, each point in the registered point clouds $\hat{\mathcal{P}}_j$ is assigned to different semantic spaces. During the training phase, the features of points belonging to each semantic space are embedded into their corresponding spaces to create normal distributions. Following this step, each semantic space $S_i$ is embedded with a normal semantic distribution, and these distributions may intersect and are not completely orthogonal. During testing, it is necessary to traverse all semantic spaces to find the nearest neighbor. Since different semantic spaces have undergone different preprocessing, their normal features tend to follow the feature distribution of their respective semantic spaces. This makes anomaly judgments more likely to be confined within spaces with the same semantics. However, this method has limitations, as interference between different semantic spaces cannot be eliminated. Our proposed Patch3D optimizes this aspect.

\subsubsection{Normalization and Standardization}
Normalization and standardization refer to processing the point cloud to the same scale, making the features comparable and aiding in better convergence of the model. Of these, normalisation is a common strategy, while standardisation is not common in 3DAD. These two approaches do not differentiate between specific semantics and perform the same preprocessing on each point cloud, treating the entire point cloud as a sample-level semantics space. Consequently, the approach is deficient in aligning semantic effects, yet it remains consistent with our Fence Theorem.

Suppose the point cloud to be processed by the normalisation and normalisation approach is: $\mathcal{P}$\text{=}$\{p_1,p_2,\ldots,p_n\}$, and its corresponding preprocessing action is: $\mathcal{A}$\text{=}$\{\mathcal{A}_1,\mathcal{A}_2,\ldots,\mathcal{A}_n\}$. This means that each point is subjected to a different normalisation and standardisation pre-processing. Suppose the post-processing point cloud $\hat{\mathcal{P}} $\text{=}$\{\hat{p}_1,\hat{p}_2,\ldots,\hat{p}_n\}$, which means that a total of $n$ points are processed and the number of points before and after the processing remains the same, only the scale is transformed.

\noindent \textbf{Goal1} Standardization maps the point cloud coordinates to a distribution with mean 0 and variance 1, and Normalization scales to a fixed range (e.g., [0,1]), eliminating scale differences between different semantic structures. This process implicitly eliminates cross-semantic coupling due to scale differences and reduces simple cross-semantic interference.

\noindent \textbf{Goal2}
Global scale unification treats the entire point cloud as a single semantic space, forcing anomaly scoring to rely on relative differences within the unified space ($s_{test}=s_{nor}$ in Eq. 3 holds constant). The semantic space ranges globally, which implies a large range of anomaly judgements, which has a large impact on the distribution of features for anomaly detection and weak constraints.

\noindent \textbf{Stage1}
The preprocessing operation $\mathcal{A}$ applies a uniform transformation to the full point cloud: $\hat{p}_i = (x_i/\sigma_x, y_i/\sigma_y, z_i/\sigma_z)$ (normalisation) or $\hat{p}_i = (x_i - x_{\min})/(x_{\max} - x_{\min })$ (normalisation). Although the semantic space is not explicitly partitioned, the global comparable space is implicitly constructed by distributional alignment, which satisfies the extreme case ($n=1$) of $\mathcal{P} = \bigsqcup_{k=1}^1 \mathcal{P}_k$ in Eq. 1, and thus makes them all belong to the same semantic space at any comparison.

\noindent \textbf{Stage2}
The feature extractor $\mathcal{F}$ models the feature distribution at a uniform scale, with the orthogonality constraint degenerating to global distributional consistency ($i=j=1$ in Eq. 2). The anomaly score simplifies to $AS = \|\mathcal{F}(\hat{p}_{test}) - \mathcal{F}(\hat{p}_{nor})\|_2$, relying on reconstruction error in a single space. The level of constraints is weak, but it is also consistent with the form of the Fence Theorem.






\subsection{More Additions to the Fence Theorem}
\label{sec2}
In this section, we add some theorems and analyse the reasons why the limitations arise, and then analyse the properties when different constraints are satisfied in ideal and real case. We conclude with suggestions for future work on 3DAD. Finally, We conclude with suggestions for future work on 3DAD.

\subsubsection{Supplementary Theorem}
We add here more fundamental theorems for anomaly detection, which come from actual experiments with empirical evidence, intuition and simple mathematical proofs for most anomaly detection. They are represented in Figure 1 and reported below:
\begin{itemize}
    \item \textbf{Semantic Invariance}: For point clouds that belong to the same class, a specific structure within the point cloud should maintain the same semantic context, regardless of preprocessing methods such as rotation, translation, and scaling, or partial deformations like protrusions or concave anomalies. This structural aspect may refer to a point, superpoint, or object.
    \item \textbf{Context Specificity}: Identifying whether a point cloud structure is anomalous requires a specific semantic context; without this context, detecting anomalies becomes meaningless.
    \item \textbf{Modeling Consistency}: Any structure that has the same semantic context should exhibit a consistent distribution of features.
\end{itemize}

\subsubsection{Reason of Limitation}
\begin{figure}[t] 
    \centering 
    \includegraphics[width=1\linewidth]{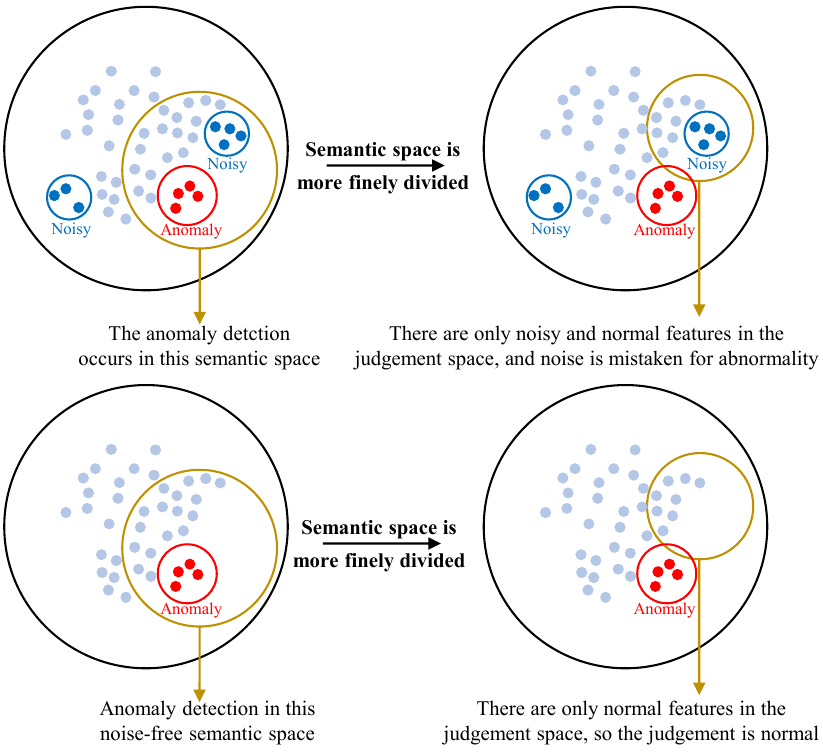} 
    \caption{Visualisation of the causes of limitations}
    \label{reason}
\end{figure}
We visualise the reasons for the existence of the limitations in Figure~\ref{reason} and analyse the full reasons. Noise itself does not have sufficient semantic representation, as demonstrated in sub-figures (a) and (b). When the semantic space used for anomaly detection is gradually reduced, the semantic division of each point becomes more precise, but this is accompanied by an increase in the influence of noise, which leads to erroneous anomaly judgements, as shown in sub-figure (b). The judgement of the noise-free space is then normal, as shown in sub-figures (c) and sub-figures (d). If the influence of noise gradually increases, the chance of misjudgement occurrence rises tremendously under the general lack of robustness of the current model.

\subsubsection{Ideal Case: Constraints Satisfied}
The preprocessing operation, denoted by $\mathcal{A}$ , must strictly satisfy the orthogonality constraint for Equation 2 to be satisfied. This requires the preprocessing approach to divide the semantic space exactly according to the semantic $s_i$ of each point $p_i$, without misclassification. Additionally, point cloud acquisition must be free of noise. The following properties hold:  
\begin{itemize}
\item \textbf{Cross-Semantic Independence}: processed subspaces $\{S_k\}$ and features $\{f_k\}$ are orthogonal to each other. 
\item \textbf{Reliable Anomaly Detection}: the anomaly score depends entirely on intra-semantic comparisons, relying on Equation 3 for scoring.
\end{itemize}

\subsubsection{Real Case: Constraints Violated}
Orthogonality is a challenging concept to realise in the context of practical anomaly detection operations. This challenge arises from the difficulty in identifying an optimal preprocessing algorithm, which hinders the accurate division of the semantic space. Additionally, the presence of noise or unpredictable perturbations during the acquisition of the point cloud cannot be eliminated. This process can be formally described as:
\begin{equation}
\exists i, j \in \{1, \ldots, n\}, i \neq j, \ \operatorname{tr}(S_i^\top S_j) \neq 0, \operatorname{tr}(f_i^\top f_j) \neq 0.
\end{equation}
The following properties hold:  
\begin{itemize}
\item \textbf{Cross-Semantic Interferencen}: When the structural matrices $S_i$ and $S_j$ of semantic subspaces are non-orthogonal, the off-diagonal terms of their covariance matrix $\Sigma_{ij}$ \text{=} $\mathbb{E}[f_i f_j^\top]$ become non-zero ($\Sigma_{ij} \neq \mathbf{0}$), indicating statistical correlations between distinct semantic features. This coupling causes overlapping anomaly regions in the feature space, making it challenging for detection algorithms to distinguish cross-semantic anomaly patterns.  

\item \textbf{Unbelievable Point Scores:} Due to cross-semantic interference, as distinguished from equation~3 in the main body of a text, the anomaly score is rewritten and decomposed into target and interference terms:  
\begin{equation}
    \mathcal{AS} \text{=} \underbrace{\|f_i^{\text{nor}} - f_i^{\text{test}}\|_2}_{\text{Target Term}} + \underbrace{\sum_{j \neq i} \alpha_{ij} \|f_j^{\text{test}}\|_2}_{\text{Interference Term}},
\end{equation}
where $\mathcal{AS}$ stands for Abnormal Score and the interference coefficient $\alpha_{ij}$ correlates with the degree of cross-semantic interference, leading to unpredictable false positives or false negatives.
\end{itemize}

\subsubsection{Suggestion for Feature Work}
Future WorkThe Fence Theorem has been comprehensively analysed, and it is understood that future research should focus on two key areas: 1) the creation of a more accurate and orthogonal semantic space during preprocessing, and 2) the enhancement of the representation of discriminative features within the semantic space.

\noindent\textbf{1) The creation of a more accurate and orthogonal semantic space during preprocessing} will be achieved by utilising the Patch-Cutting and Patch-Matching approaches to create a semantic space, relying on K-Means clustering and the optimisation process of similar semantic points to merge the spaces.However, challenges were encountered when dealing with deformed datasets. We propose a more efficient approach that is more precise, robust and with Rotationally Invariant Feature Approach to accurately segment each point into its correct semantic space, in addition to guaranteeing that points within different semantic spaces that need to be different do not affect each other.

\noindent\textbf{2) The enhancement of the representation of discriminative features within the semantic space.} Our goal is to achieve a more discriminative representation for each point. In this study, we adopt the FPFH feature as the exact descriptor for each point due to its mathematical interpretability and its advantages in terms of both computational speed and accuracy.However, traditional feature descriptors have their limitations.Therefore, we need more discriminative features that increase the gap between normal and anomalous features in the semantic space, making it easier for anomalous features to deviate from this normal distribution.
\subsection{More Results and Analyses of Experimental Data}
\label{sec3}
We provide in this section the complete data we used in the experimental part, including the measured data of Patch3D as well as the measured data of existing approaches.

\subsubsection{Registration Approach}
We provide experimental data measuring PatchCore (FPFH+Raw), Reg3D-AD and including O-AUROC, O-AUPR, P-AUROC, P-AUPR, and FPS, which are presented in Table~\ref{L1}, \ref{L2}, \ref{L3}, \ref{L4} and \ref{TIME}, respectively, to confirm our conclusions. The anomaly detection accuracy gradually improves as the registration accuracy increases, as indicated by the consistency of the visual content representation in the body, due to the fact that as the registration accuracy increases, each point is progressively classified into the correct semantic space and receives the correct preprocessing, increasing the comparability of point-level features with the same semantics, and better restricting the discriminative process of anomaly detection to the same semantics. Although the feature extraction approaches are different, e.g. PatchCore (FPFH+Raw) utilises FPFH features with coordinates, while Reg3D-AD utilises PointMAE features and coordinates, and ISMP utilises global features from pseudo-modalities, PointMAE features, and FPFH features, which brings about a huge gap in feature extraction, they both use the the same RANSAC registration approach, which brings comparability. The results show the unity of the conclusions despite the huge difference in feature extraction capabilities. This is similar to the creation of pseudo-anomalies approach, where different approaches to creating pseudo-anomalies are unified to the same conclusion: as the semantic space between anomalous and normal is partitioned more explicitly, and each semantic space is processed more correctly, anomaly detection becomes better. In addition, the code for RANSAC is the same as Reg3DAD, and the variable R modified in our paper controls the voxel size, which further controls the registration accuracy; the smaller R is the higher the registration accuracy.

\subsubsection{Creating Pseudo-Anomalies}
The experimental data for creating pseudo-anomalous approaches are presented in Tables 1, 2, 3 and 4, respectively.


\subsubsection{Patch3D}
\begin{figure}[t] 
    \centering 
    \includegraphics[width=\linewidth]{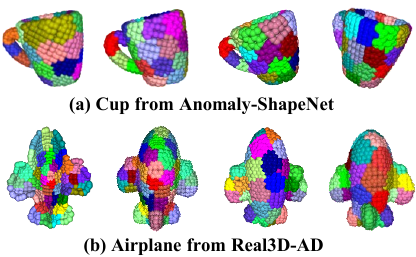} 
    \caption{Simple visualisation of the semantic space delineated by Patch3D. }
    \label{more}
\end{figure}
We provide the complete data used in the experimental part of the main text in this section, in addition, we provide a simple One-Shot experiment to further describe the anomaly detection effect, this is due to the fact that One-Shot is more of a test of the anomaly detection ability of the anomaly detection model. Furthermore, the semantic space delineated by Patch3D is simply visualised in Figure~\ref{more}, which is the semantic space created after Patch-Cutting. The complete O-AUROC, O-AUPR, P-AUROC and P-AUPR values are shown in Tables~\ref{L5}, \ref{L6}, \ref{L7} and \ref{L8}, respectively, along with the values and their means for each class in turn. One-Shot experiments were briefly added here, and P-AUROC and O-AUROC are shown in Table~\ref{13} and \ref{14}, respectively. We can observe that compared to the BTF, which only divides one semantic space, we chose to divide 40 semantic spaces, and the P-AUROC is improved by 20.9\%, which means that the point-level detection capability is greatly improved.

\begin{table*}[thbp]
\centering
\resizebox{0.95\textwidth}{!}{
\begin{tabular}{c|ccccccccccccc|cc}
    \toprule
    \multicolumn{16}{c}{\textbf{O-AUROC}} \\
    \hline
    \textbf{Method} & \textit{\textbf{R}} & \textbf{airplane} & \textbf{car} & \textbf{candybar} & \textbf{chicken} & \textbf{diamond} & \textbf{duck} & \textbf{fish} & \textbf{gemstone} & \textbf{seahorse} & \textbf{shell} & \textbf{starfish} & \textbf{Toffees} & \textbf{Mean} & \textbf{Bias} \\
    \hline
    \multirow{11}[1]{*}{\textbf{Reg3D-AD}} & 0.80  & 0.538  & 0.604  & 0.641  & 0.547  & 0.708  & 0.600  & 0.616  & 0.479  & 0.566  & 0.492  & 0.476  & 0.675  & 0.579  & 0.001  \\
          & 0.75  & 0.612  & 0.670  & 0.610  & 0.583  & 0.752  & 0.588  & 0.699  & 0.487  & 0.628  & 0.587  & 0.534  & 0.705  & 0.621  & 0.007  \\
          & 0.70  & 0.593  & 0.649  & 0.745  & 0.607  & 0.767  & 0.564  & 0.704  & 0.445  & 0.559  & 0.637  & 0.590  & 0.657  & 0.626  & 0.007  \\
          & 0.65  & 0.577  & 0.620  & 0.648  & 0.640  & 0.776  & 0.612  & 0.777  & 0.483  & 0.607  & 0.502  & 0.579  & 0.622  & 0.621  & 0.002  \\
          & 0.60  & 0.597  & 0.707  & 0.794  & 0.649  & 0.809  & 0.590  & 0.794  & 0.458  & 0.718  & 0.553  & 0.580  & 0.667  & 0.660  & 0.007  \\
          & 0.55  & 0.622  & 0.663  & 0.764  & 0.619  & 0.859  & 0.567  & 0.882  & 0.502  & 0.719  & 0.572  & 0.597  & 0.707  & 0.673  & 0.011  \\
\hline
    \end{tabular}%
}
\caption{O-AUROC results of exiting approach reviews. The bias represents the distance from the mean of the data point that is the furthest point from the mean after multiple experiments.
 All experiments were done on two RTX 2080Ti.\label{L1}}
\end{table*}

\begin{table*}[htbp]
\centering
\resizebox{0.95\textwidth}{!}{
     \begin{tabular}{c|ccccccccccccc|cc}
    \hline
    \multicolumn{16}{c}{\textbf{O-AUPR}} \\
    \hline
    \textbf{Method} & \textit{\textbf{R}} & \textbf{airplane} & \textbf{car} & \textbf{candybar} & \textbf{chicken} & \textbf{diamond} & \textbf{duck} & \textbf{fish} & \textbf{gemstone} & \textbf{seahorse} & \textbf{shell} & \textbf{starfish} & \textbf{Toffees} & \textbf{Mean} & \textbf{Bias} \\
    \hline
    \multirow{11}[2]{*}{\textbf{Reg3D-AD}} & 0.80  & 0.517  & 0.600  & 0.615  & 0.539  & 0.703  & 0.587  & 0.678  & 0.504  & 0.608  & 0.557  & 0.504  & 0.677  & 0.591  & 0.001  \\
          & 0.75  & 0.629  & 0.657  & 0.613  & 0.569  & 0.680  & 0.567  & 0.735  & 0.509  & 0.638  & 0.569  & 0.554  & 0.717  & 0.620  & 0.004  \\
          & 0.70  & 0.567  & 0.628  & 0.739  & 0.592  & 0.703  & 0.549  & 0.739  & 0.473  & 0.572  & 0.637  & 0.619  & 0.684  & 0.625  & 0.005  \\
          & 0.65  & 0.581  & 0.621  & 0.641  & 0.611  & 0.733  & 0.593  & 0.818  & 0.511  & 0.657  & 0.548  & 0.567  & 0.671  & 0.629  & 0.004  \\
          & 0.60  & 0.620  & 0.703  & 0.796  & 0.599  & 0.758  & 0.631  & 0.830  & 0.487  & 0.710  & 0.619  & 0.580  & 0.693  & 0.669  & 0.002  \\
          & 0.55  & 0.635  & 0.649  & 0.774  & 0.585  & 0.796  & 0.552  & 0.902  & 0.495  & 0.742  & 0.657  & 0.565  & 0.707  & 0.672  & 0.008  \\
\hline
    \end{tabular}%
}
\caption{O-AUPR results of exiting approach reviews. The bias represents the distance from the mean of the data point that is the furthest point from the mean after multiple experiments.
 All experiments were done on two RTX 2080Ti.\label{L2}}
\end{table*}

\begin{table*}[htbp]
\centering
\resizebox{0.95\textwidth}{!}{
    \begin{tabular}{c|ccccccccccccc|cc}
    \toprule
    \multicolumn{16}{c}{\textbf{P-AUROC}} \\
    \hline
    \textbf{Method} & \textit{\textbf{R}} & \textbf{airplane} & \textbf{car} & \textbf{candybar} & \textbf{chicken} & \textbf{diamond} & \textbf{duck} & \textbf{fish} & \textbf{gemstone} & \textbf{seahorse} & \textbf{shell} & \textbf{starfish} & \textbf{Toffees} & \textbf{Mean} & \textbf{Bias} \\
    \hline
    \multirow{11}[2]{*}{\textbf{Reg3D-AD}} & 0.80  & 0.537  & 0.715  & 0.710  & 0.676  & 0.680  & 0.524  & 0.713  & 0.488  & 0.662  & 0.687  & 0.560  & 0.798  & 0.646 & 0.006 \\
          & 0.75  & 0.571  & 0.729  & 0.698  & 0.684  & 0.694  & 0.526  & 0.743  & 0.505  & 0.711  & 0.729  & 0.563  & 0.796  & 0.663 & 0.005 \\
          & 0.70  & 0.591  & 0.722  & 0.729  & 0.686  & 0.734  & 0.533  & 0.780  & 0.476  & 0.695  & 0.721  & 0.563  & 0.795  & 0.669 & 0.005 \\
          & 0.65  & 0.590  & 0.714  & 0.702  & 0.705  & 0.692  & 0.514  & 0.789  & 0.518  & 0.684  & 0.731  & 0.578  & 0.784  & 0.6665 & 0.0005 \\
          & 0.60  & 0.576  & 0.719  & 0.736  & 0.703  & 0.747  & 0.657  & 0.799  & 0.521  & 0.763  & 0.746  & 0.589  & 0.787  & 0.695 & 0.005 \\
          & 0.55  & 0.596  & 0.716  & 0.701  & 0.695  & 0.767  & 0.517  & 0.807  & 0.489  & 0.778  & 0.755  & 0.579  & 0.795  & 0.683 & 0.003 \\
\hline
    \end{tabular}%
}
\caption{P-AUROC results of exiting approach reviews. The bias represents the distance from the mean of the data point that is the furthest point from the mean after multiple experiments.
 All experiments were done on two RTX 2080Ti.\label{L3}}
\end{table*}

\begin{table*}[htbp]
\centering
\resizebox{0.95\textwidth}{!}{
    \begin{tabular}{c|ccccccccccccc|cc}
    \hline
    \multicolumn{16}{c}{\textbf{P-AUPR}} \\
    \hline
    \textbf{Method} & \textit{\textbf{R}} & \textbf{airplane} & \textbf{car} & \textbf{candybar} & \textbf{chicken} & \textbf{diamond} & \textbf{duck} & \textbf{fish} & \textbf{gemstone} & \textbf{seahorse} & \textbf{shell} & \textbf{starfish} & \textbf{Toffees} & \textbf{Mean} & \textbf{Bias} \\
    \hline
    \multirow{11}[2]{*}{\textbf{Reg3D-AD}} & 0.80  & 0.008  & 0.084  & 0.042  & 0.040  & 0.055  & 0.011  & 0.147  & 0.007  & 0.048  & 0.020  & 0.037  & 0.076  & 0.048 & 0.003 \\
          & 0.75  & 0.010  & 0.136  & 0.030  & 0.042  & 0.067  & 0.011  & 0.170  & 0.008  & 0.069  & 0.026  & 0.035  & 0.071  & 0.056 & 0.002 \\
          & 0.70  & 0.011  & 0.116  & 0.063  & 0.046  & 0.083  & 0.012  & 0.228  & 0.007  & 0.068  & 0.023  & 0.041  & 0.075  & 0.064 & 0.004 \\
          & 0.65  & 0.010  & 0.112  & 0.051  & 0.052  & 0.068  & 0.011  & 0.308  & 0.008  & 0.066  & 0.034  & 0.060  & 0.076  & 0.0715 & 0.0005 \\
          & 0.60  & 0.010  & 0.134  & 0.080  & 0.053  & 0.086  & 0.092  & 0.347  & 0.008  & 0.171  & 0.044  & 0.040  & 0.074  & 0.095 & 0.004 \\
          & 0.55  & 0.011  & 0.112  & 0.084  & 0.057  & 0.086  & 0.010  & 0.396  & 0.008  & 0.159  & 0.049  & 0.042  & 0.072  & 0.091 & 0.003 \\
\hline
    \end{tabular}%
}
\caption{P-AUPR results of exiting approach reviews. The bias represents the distance from the mean of the data point that is the furthest point from the mean after multiple experiments.
 All experiments were done on two RTX 2080Ti.\label{L4}}
\end{table*}

\begin{table*}[htbp]
\centering
\resizebox{0.95\textwidth}{!}{
    \begin{tabular}{c|ccccccccccccc|cc}
    \hline
    \multicolumn{16}{c}{\textbf{FPS}} \\
    \hline
    \textbf{Method} & \textit{\textbf{R}} & \textbf{airplane} & \textbf{car} & \textbf{candybar} & \textbf{chicken} & \textbf{diamond} & \textbf{duck} & \textbf{fish} & \textbf{gemstone} & \textbf{seahorse} & \textbf{shell} & \textbf{starfish} & \textbf{Toffees} & \textbf{Mean} & \textbf{Bias} \\
    \hline
    \multirow{11}[2]{*}{\textbf{Reg3D-AD}} & 0.80  & 0.133  & 0.204  & 0.293  & 0.123  & 0.206  & 0.134  & 0.274  & 0.242  & 0.311  & 0.246  & 0.298  & 0.212  & 0.202  & 0.00032 \\
          & 0.75  & 0.132  & 0.202  & 0.289  & 0.122  & 0.206  & 0.131  & 0.270  & 0.243  & 0.311  & 0.245  & 0.299  & 0.213  & 0.201  & 0.00004 \\
          & 0.70  & 0.132  & 0.201  & 0.285  & 0.121  & 0.205  & 0.131  & 0.264  & 0.241  & 0.311  & 0.244  & 0.298  & 0.212  & 0.199  & 0.00004 \\
          & 0.65  & 0.129  & 0.201  & 0.283  & 0.121  & 0.201  & 0.128  & 0.265  & 0.240  & 0.306  & 0.239  & 0.292  & 0.212  & 0.197  & 0.00004 \\
          & 0.60  & 0.128  & 0.201  & 0.278  & 0.120  & 0.202  & 0.177  & 0.263  & 0.237  & 0.308  & 0.239  & 0.297  & 0.209  & 0.204  & 0.0088 \\
          & 0.55  & 0.123  & 0.192  & 0.276  & 0.119  & 0.198  & 0.123  & 0.263  & 0.232  & 0.303  & 0.233  & 0.289  & 0.206  & 0.192  & 0.00022 \\
\hline
    \end{tabular}%
}
\caption{FPS results of exiting approach reviews. The bias represents the distance from the mean of the data point that is the furthest point from the mean after multiple experiments.
 All experiments were done on two RTX 2080Ti.\label{TIME}}
\end{table*}

\begin{table*}[htbp]
\centering
\resizebox{0.95\textwidth}{!}{
\begin{tabular}{c|cccccccccccc|cc}
    \hline
    \multicolumn{15}{c}{\textbf{O-AUROC}} \\
    \hline
    \textbf{Method} & \textbf{ashtray0} & \textbf{bag0} & \textbf{bottle0} & \textbf{bottle1} & \textbf{bottle3} & \textbf{bowl0} & \textbf{bowl1} & \textbf{bowl2} & \textbf{bowl3} & \textbf{bowl4} & \textbf{bowl5} & \multicolumn{1}{c}{\textbf{bucket0}} & \textbf{bucket1} & \textbf{cap0} \\
    \hline
    \textbf{Patch3D\_1} & 0.4905  & 0.5524  & 0.3714  & 0.4491  & 0.5365  & 0.9222  & 0.6852  & 0.4963  & 0.4482  & 0.5778  & 0.3895  & \multicolumn{1}{c}{0.8397 } & 0.6222  & 0.8444  \\
    \textbf{Patch3D\_2} & 0.5286  & 0.5048  & 0.5143  & 0.3667  & 0.6889  & 0.5852  & 0.7963  & 0.4296  & 0.4074  & 0.8482  & 0.5790  & \multicolumn{1}{c}{0.8238 } & 0.7016  & 0.8148  \\
    \textbf{Patch3D\_3} & 0.4667  & 0.4905  & 0.4286  & 0.5895  & 0.6825  & 0.4889  & 0.6667  & 0.4444  & 0.6778  & 0.8482  & 0.4281  & \multicolumn{1}{c}{0.7937 } & 0.6413  & 0.7222  \\
    \textbf{Patch3D\_4} & 0.5429  & 0.5381  & 0.5381  & 0.3088  & 0.6635  & 0.5074  & 0.5333  & 0.4000  & 0.5185  & 0.8778  & 0.6667  & \multicolumn{1}{c}{0.7302 } & 0.6254  & 0.7556  \\
    \textbf{Patch3D\_5} & 0.4905  & 0.4524  & 0.5381  & 0.3263  & 0.5524  & 0.6000  & 0.6556  & 0.6074  & 0.5519  & 0.5815  & 0.4842  & \multicolumn{1}{c}{0.7492 } & 0.5873  & 0.7667  \\
    \textbf{Patch3D\_6} & 0.6191  & 0.7333  & 0.6143  & 0.3158  & 0.6222  & 0.7185  & 0.4778  & 0.7148  & 0.2778  & 0.3815  & 0.4702  & \multicolumn{1}{c}{0.7492 } & 0.6222  & 0.6074  \\
\hline
    \end{tabular}%

}
\caption{O-AUROC results for Patch3D review on Anomaly-ShapeNet. All experiments were done on two RTX 2080Ti.\label{L5}}
\end{table*}

\begin{table*}[htbp]
\centering
\resizebox{0.95\textwidth}{!}{
    \begin{tabular}{c|cccccccccccc|cc}
    \hline
    \multicolumn{15}{c}{\textbf{O-AUPR}} \\
    \hline
    \textbf{Method} & \textbf{ashtray0} & \textbf{bag0} & \textbf{bottle0} & \textbf{bottle1} & \textbf{bottle3} & \textbf{bowl0} & \textbf{bowl1} & \textbf{bowl2} & \textbf{bowl3} & \textbf{bowl4} & \textbf{bowl5} & \multicolumn{1}{c}{\textbf{bucket0}} & \textbf{bucket1} & \textbf{cap0} \\
    \hline
    \textbf{Patch3D\_1} & 0.5054  & 0.5532  & 0.3926  & 0.5372  & 0.6168  & 0.9545  & 0.7132  & 0.4898  & 0.4785  & 0.6384  & 0.4582  & \multicolumn{1}{c}{0.9106 } & 0.6279  & 0.8638  \\
    \textbf{Patch3D\_2} & 0.5338  & 0.4777  & 0.5058  & 0.4731  & 0.7337  & 0.6343  & 0.7945  & 0.4588  & 0.4554  & 0.8748  & 0.5739  & \multicolumn{1}{c}{0.8987 } & 0.8030  & 0.8463  \\
    \textbf{Patch3D\_3} & 0.4550  & 0.5293  & 0.4262  & 0.6647  & 0.7641  & 0.5967  & 0.6767  & 0.4794  & 0.6291  & 0.8726  & 0.4799  & \multicolumn{1}{c}{0.8665 } & 0.7168  & 0.6960  \\
    \textbf{Patch3D\_4} & 0.5501  & 0.5525  & 0.4684  & 0.4397  & 0.7739  & 0.5340  & 0.6212  & 0.4488  & 0.5315  & 0.8918  & 0.7781  & \multicolumn{1}{c}{0.8596 } & 0.6655  & 0.7960  \\
    \textbf{Patch3D\_5} & 0.5816  & 0.7112  & 0.7124  & 0.6258  & 0.7440  & 0.7360  & 0.8441  & 0.8942  & 0.7986  & 0.8869  & 0.4024  & \multicolumn{1}{c}{0.7304 } & 0.8195  & 0.8717  \\
    \textbf{Patch3D\_6} & 0.6143  & 0.7903  & 0.6706  & 0.5075  & 0.6402  & 0.7319  & 0.5185  & 0.7506  & 0.4051  & 0.5087  & 0.4994  & \multicolumn{1}{c}{0.8732 } & 0.6900  & 0.6320  \\
\hline
    \end{tabular}%
}
\caption{O-AUPR results for Patch3D review on Anomaly-ShapeNet. All experiments were done on two RTX 2080Ti.\label{L6}}

\end{table*}

\begin{table*}[htbp]
\centering
\resizebox{0.95\textwidth}{!}{
 \begin{tabular}{c|cccccccccccc|cc}
    \hline
    \multicolumn{15}{c}{\textbf{P-AUROC}} \\
    \hline
    \textbf{Method} & \textbf{ashtray0} & \textbf{bag0} & \textbf{bottle0} & \textbf{bottle1} & \textbf{bottle3} & \textbf{bowl0} & \textbf{bowl1} & \textbf{bowl2} & \textbf{bowl3} & \textbf{bowl4} & \textbf{bowl5} & \multicolumn{1}{c}{\textbf{bucket0}} & \textbf{bucket1} & \textbf{cap0} \\
    \hline
    \textbf{Patch3D\_1} & 0.6022  & 0.6960  & 0.7073  & 0.6031  & 0.5423  & 0.7104  & 0.8440  & 0.9121  & 0.8100  & 0.8770  & 0.3544  & \multicolumn{1}{c}{0.7306 } & 0.8357  & 0.8428  \\
    \textbf{Patch3D\_2} & 0.5629  & 0.6485  & 0.6855  & 0.6362  & 0.6088  & 0.7626  & 0.8506  & 0.9156  & 0.8451  & 0.8852  & 0.4118  & \multicolumn{1}{c}{0.7305 } & 0.8323  & 0.8469  \\
    \textbf{Patch3D\_3} & 0.5895  & 0.6787  & 0.7090  & 0.5973  & 0.5501  & 0.7830  & 0.8681  & 0.9023  & 0.7968  & 0.8880  & 0.4815  & \multicolumn{1}{c}{0.7399 } & 0.8259  & 0.8523  \\
    \textbf{Patch3D\_4} & 0.6166  & 0.6454  & 0.6931  & 0.5739  & 0.6548  & 0.8064  & 0.8417  & 0.9097  & 0.8382  & 0.8986  & 0.3870  & \multicolumn{1}{c}{0.7385 } & 0.8201  & 0.8538  \\
    \textbf{Patch3D\_5} & 0.5066  & 0.4793  & 0.5470  & 0.4533  & 0.5750  & 0.6510  & 0.6774  & 0.6186  & 0.6369  & 0.5871  & 0.6560  & \multicolumn{1}{c}{0.8331 } & 0.6503  & 0.7481  \\
    \textbf{Patch3D\_6} & 0.5616  & 0.6500  & 0.6501  & 0.5845  & 0.7668  & 0.7814  & 0.8135  & 0.9104  & 0.7971  & 0.8818  & 0.4883  & \multicolumn{1}{c}{0.7390 } & 0.8199  & 0.8676  \\
\hline
    \end{tabular}%
}
\caption{P-AUROC results for Patch3D review on Anomaly-ShapeNet. All experiments were done on two RTX 2080Ti.\label{L7}}
\end{table*}

\begin{table*}[htbp]
\centering
\resizebox{0.95\textwidth}{!}{
    \begin{tabular}{c|cccccccccccc|cc}
    \hline
    \multicolumn{15}{c}{P-AUPR} \\
    \hline
    \textbf{Method} & \textbf{ashtray0} & \textbf{bag0} & \textbf{bottle0} & \textbf{bottle1} & \textbf{bottle3} & \textbf{bowl0} & \textbf{bowl1} & \textbf{bowl2} & \textbf{bowl3} & \textbf{bowl4} & \textbf{bowl5} & \multicolumn{1}{c}{\textbf{bucket0}} & \textbf{bucket1} & \textbf{cap0} \\
    \hline
    \textbf{Patch3D\_1} & 0.0625  & 0.0159  & 0.0281  & 0.0201  & 0.0111  & 0.0607  & 0.0338  & 0.1174  & 0.0490  & 0.0733  & 0.0066  & \multicolumn{1}{c}{0.0438 } & 0.0873  & 0.1596  \\
    \textbf{Patch3D\_2} & 0.0495  & 0.0136  & 0.0276  & 0.0215  & 0.0133  & 0.0734  & 0.0354  & 0.1167  & 0.0689  & 0.0645  & 0.0072  & \multicolumn{1}{c}{0.0425 } & 0.0863  & 0.1586  \\
    \textbf{Patch3D\_3} & 0.0575  & 0.0150  & 0.0267  & 0.0224  & 0.0114  & 0.0749  & 0.0468  & 0.1126  & 0.0187  & 0.0705  & 0.0082  & \multicolumn{1}{c}{0.0467 } & 0.0751  & 0.1545  \\
    \textbf{Patch3D\_4} & 0.0639  & 0.0138  & 0.0264  & 0.0174  & 0.0169  & 0.0981  & 0.0326  & 0.1149  & 0.0335  & 0.0750  & 0.0069  & \multicolumn{1}{c}{0.0452 } & 0.0677  & 0.1649  \\
    \textbf{Patch3D\_5} & 0.0561  & 0.0171  & 0.0291  & 0.0213  & 0.0315  & 0.0551  & 0.0367  & 0.0941  & 0.0182  & 0.0588  & 0.0071  & \multicolumn{1}{c}{0.0455 } & 0.0625  & 0.1602  \\
    \textbf{Patch3D\_6} & 0.0534  & 0.0136  & 0.0213  & 0.0190  & 0.0421  & 0.0759  & 0.0195  & 0.0806  & 0.0189  & 0.0420  & 0.0083  & \multicolumn{1}{c}{0.0456 } & 0.0655  & 0.1360  \\
\hline
    \end{tabular}%
}
\caption{P-AUPR results for Patch3D review on Anomaly-ShapeNet. All experiments were done on two RTX 2080Ti.\label{L8}}
\end{table*}

\begin{table*}[htbp]
\centering
\resizebox{\textwidth}{!}{
    \begin{tabular}{c|cccccccccccc|c}
    \multicolumn{1}{c}{} &       &       &       &       &       &       &       &       &       &       &       & \multicolumn{1}{c}{} &  \\
    \hline
    \multicolumn{14}{c}{\textbf{O-AUROC}} \\
    \hline
    \textbf{Method} & \textbf{Airplane} & \textbf{Car} & \textbf{Candybar} & \textbf{Chicken} & \textbf{Diamond} & \textbf{Duck} & \textbf{Fish} & \textbf{Gemstone} & \textbf{Seahorse} & \textbf{Shell} & \textbf{Starfish} & \textbf{Toffees} & \textbf{Mean} \\
    \hline
    \textbf{Patch3D (1)} & 0.8815 & 0.5901  & 0.5645  & 0.8372  & 0.5736  & 0.5458  & 0.6747  & 0.3703  & 0.5050  & 0.5849  & 0.4408  & 0.5411  & 0.5925  \\
    \textbf{Patch3D (2)} & 0.7621  & 0.5722  & 0.5682  & 0.7517  & 0.5732  & 0.5284  & 0.6537  & 0.4824  & 0.5121  & 0.5909  & 0.5121  & 0.7210  & 0.6023  \\
    \textbf{Patch3D (3)} & 0.7021  & 0.5472  & 0.5868  & 0.7017  & 0.5746  & 0.5358  & 0.6637  & 0.5012  & 0.5333  & 0.6011  & 0.5843  & 0.6892  & 0.6018  \\
    \textbf{Patch3D (4)} & 0.6202  & 0.5388  & 0.5615  & 0.6302  & 0.5884  & 0.5104  & 0.6314  & 0.4355  & 0.5642  & 0.5921  & 0.5433  & 0.7518  & 0.5807  \\
    \textbf{Patch3D (5)} & 0.6935  & 0.5328  & 0.5913  & 0.5581  & 0.5834  & 0.5382  & 0.6448  & 0.5326  & 0.5310  & 0.5901  & 0.5728  & 0.7121  & 0.5901  \\
\hline
    \end{tabular}%
}
\caption{O-AUROC results for Patch3D review on Real3D-AD. In the absence of a comprehensive deep learning process for all computations, there is an absence of repetitive bias. All experiments were done on two RTX 2080Ti.\label{L9}}
\end{table*}

\begin{table*}[htbp]
\centering
\resizebox{\textwidth}{!}{
    \begin{tabular}{c|cccccccccccc|c}
    \hline
    \multicolumn{14}{c}{\textbf{O-AUPR}} \\
    \hline
    \textbf{Method} & \textbf{Airplane} & \textbf{Car} & \textbf{Candybar} & \textbf{Chicken} & \textbf{Diamond} & \textbf{Duck} & \textbf{Fish} & \textbf{Gemstone} & \textbf{Seahorse} & \textbf{Shell} & \textbf{Starfish} & \textbf{Toffees} & \textbf{Mean} \\
    \hline
    \textbf{Patch3D (1)} & \textcolor[rgb]{ .024,  .024,  .027}{0.8715 } & 0.5951  & 0.5685  & 0.8461  & 0.5801  & 0.5494  & 0.6646  & 0.3742  & 0.5063  & 0.5969  & 0.4434  & 0.5400  & 0.5947  \\
    \textbf{Patch3D (2)} & \textcolor[rgb]{ .024,  .024,  .027}{0.7710 } & 0.5853  & 0.5694  & 0.7646  & 0.5728  & 0.5296  & 0.6606  & 0.4893  & 0.5133  & 0.6026  & 0.5144  & 0.7263  & 0.6083  \\
    \textbf{Patch3D (3)} & 0.7041  & 0.5132  & 0.5854  & 0.7100  & 0.5832  & 0.5356  & 0.6702  & 0.5046  & 0.5217  & 0.6127  & 0.5962  & 0.6884  & 0.6021  \\
    \textbf{Patch3D (4)} & 0.6411  & 0.5297  & 0.5579  & 0.6421  & 0.5904  & 0.5173  & 0.6243  & 0.4482  & 0.5758  & 0.6122  & 0.5677  & 0.7654  & 0.5893  \\
    \textbf{Patch3D (5)} & 0.6935  & 0.5328  & 0.5913  & 0.5581  & 0.5834  & 0.5382  & 0.6448  & 0.5326  & 0.5310  & 0.5901  & 0.5728  & 0.7121  & 0.5901  \\
    \textbf{Patch3D (6)} & 0.5571  & 0.5601  & 0.5879  & 0.6043  & 0.5901  & 0.5124  & 0.6327  & 0.5139  & 0.5419  & 0.5834  & 0.5624  & 0.7427  & 0.5824  \\
\hline
    \end{tabular}%
}
\caption{O-AUPR results for Patch3D review on Real3D-AD. In the absence of a comprehensive deep learning process for all computations, there is an absence of repetitive bias. All experiments were done on two RTX 2080Ti.\label{10}}
\end{table*}

\begin{table*}[htbp]
\centering
\resizebox{\textwidth}{!}{
    \begin{tabular}{c|cccccccccccc|c}
    \hline
    \multicolumn{14}{c}{\textbf{P-AUROC}} \\
    \hline
    \textbf{Method} & \textbf{Airplane} & \textbf{Car} & \textbf{Candybar} & \textbf{Chicken} & \textbf{Diamond} & \textbf{Duck} & \textbf{Fish} & \textbf{Gemstone} & \textbf{Seahorse} & \textbf{Shell} & \textbf{Starfish} & \textbf{Toffees} & \textbf{Mean} \\
    \hline
    \textbf{Patch3D (1)} & 0.4712 & 0.6428  & 0.6374  & 0.6180  & 0.7599  & 0.4301  & 0.4641  & 0.8301  & 0.5441  & 0.5958  & 0.5224  & 0.4107  & 0.5772  \\
    \textbf{Patch3D (2)} & 0.5447  & 0.6063  & 0.6727  & 0.7000  & 0.7812  & 0.4622  & 0.5081  & 0.7421  & 0.5219  & 0.5677  & 0.5126  & 0.4221  & 0.5868  \\
    \textbf{Patch3D (3)} & 0.5356  & 0.6105  & 0.6632  & 0.7143  & 0.7739  & 0.4773  & 0.4366  & 0.7672  & 0.5228  & 0.5342  & 0.5245  & 0.4443  & 0.5837  \\
    \textbf{Patch3D (4)} & 0.4726  & 0.6230  & 0.6730  & 0.6844  & 0.7537  & 0.5601  & 0.5210  & 0.8256  & 0.5384  & 0.5665  & 0.5320  & 0.4344  & 0.5987  \\
    \textbf{Patch3D (5)} & 0.5240  & 0.6422  & 0.6611  & 0.6990  & 0.7466  & 0.5640  & 0.5394  & 0.7760  & 0.5420  & 0.5433  & 0.5580  & 0.4457  & 0.6034  \\
    \textbf{Patch3D (6)} & 0.5501  & 0.6360  & 0.6540  & 0.6894  & 0.7668  & 0.5720  & 0.5549  & 0.8032  & 0.5319  & 0.5442  & 0.5430  & 0.4349  & 0.6067  \\
\hline
    \end{tabular}%
}
\caption{P-AUROC results for Patch3D review on Real3D-AD. In the absence of a comprehensive deep learning process for all computations, there is an absence of repetitive bias. All experiments were done on two RTX 2080Ti.\label{11}}
\end{table*}

\begin{table*}[htbp]
\centering
\resizebox{\textwidth}{!}{
    \begin{tabular}{c|cccccccccccc|c}
    \hline
    \multicolumn{14}{c}{\textbf{P-AUPR}} \\
    \hline
    \textbf{Method} & \textbf{Airplane} & \textbf{Car} & \textbf{Candybar} & \textbf{Chicken} & \textbf{Diamond} & \textbf{Duck} & \textbf{Fish} & \textbf{Gemstone} & \textbf{Seahorse} & \textbf{Shell} & \textbf{Starfish} & \textbf{Toffees} & \textbf{Mean} \\
    \hline
    \textbf{Patch3D (1)} & 0.0074  & 0.0282  & 0.1182  & 0.0444  & 0.2392  & 0.0683  & 0.0361  & 0.0754  & 0.0276  & 0.0180  & 0.0348  & 0.0553  & 0.0640  \\
    \textbf{Patch3D (2)} & 0.0081  & 0.0294  & 0.1194  & 0.0471  & 0.2401  & 0.0699  & 0.0375  & 0.0763  & 0.0284  & 0.0192  & 0.0349  & 0.0561  & 0.0639  \\
    \textbf{Patch3D (3)} & 0.0090  & 0.0310  & 0.1128  & 0.0461  & 0.2349  & 0.0714  & 0.0402  & 0.0743  & 0.0277  & 0.0188  & 0.0423  & 0.0601  & 0.0641  \\
    \textbf{Patch3D (4)} & 0.0092  & 0.0292  & 0.1143  & 0.0476  & 0.2410  & 0.0702  & 0.0433  & 0.0751  & 0.0279  & 0.0190  & 0.0358  & 0.0582  & 0.0642  \\
    \textbf{Patch3D (5)} & 0.0104  & 0.0327  & 0.1157  & 0.0462  & 0.2455  & 0.0725  & 0.0399  & 0.0753  & 0.0283  & 0.0177  & 0.0358  & 0.0524  & 0.0644  \\
    \textbf{Patch3D (6)} & 0.0103  & 0.0304  & 0.1243  & 0.0488  & 0.2411  & 0.0711  & 0.0366  & 0.0724  & 0.0264  & 0.0199  & 0.0402  & 0.0514  & 0.0644  \\
\hline
    \end{tabular}%
}
\caption{P-AUPR results for Patch3D review on Real3D-AD. In the absence of a comprehensive deep learning process for all computations, there is an absence of repetitive bias. All experiments were done on two RTX 2080Ti.\label{12}}
\end{table*}

\begin{table*}[htbp]
\centering
\resizebox{\textwidth}{!}{
    \begin{tabular}{c|cccccccccccc|cc}
    \hline
    \multicolumn{15}{c}{\textbf{P-AUROC}} \\
    \hline
    \textbf{Method} & \textbf{ashtray0} & \textbf{bag0} & \textbf{bottle0} & \textbf{bottle1} & \textbf{bottle3} & \textbf{bowl0} & \textbf{bowl1} & \textbf{bowl2} & \textbf{bowl3} & \textbf{bowl4} & \textbf{bowl5} & \multicolumn{1}{c}{\textbf{bucket0}} & \textbf{bucket1} & \textbf{cap0} \\
    \hline
    \textbf{Patchcore (FPFH+Raw)} & 0.527  & 0.757  & 0.646  & 0.594  & 0.750  & 0.831  & 0.703  & 0.650  & 0.923  & 0.785  & 0.675  & \multicolumn{1}{c}{0.580 } & 0.777  & 0.918  \\
    \textbf{RegAD} & 0.611  & 0.690  & 0.606  & 0.473  & 0.745  & 0.685  & 0.492  & 0.570  & 0.625  & 0.518  & 0.453  & \multicolumn{1}{c}{0.511 } & 0.602  & 0.608  \\
    \textbf{PatchCore (Raw)} & 0.562  & 0.664  & 0.622  & 0.498  & 0.764  & 0.653  & 0.534  & 0.516  & 0.631  & 0.565  & 0.432  & \multicolumn{1}{c}{0.461 } & 0.593  & 0.631  \\
    \textbf{M3DM} & 0.638  & 0.589  & 0.623  & 0.611  & 0.778  & 0.561  & 0.503  & 0.488  & 0.557  & 0.545  & 0.414  & \multicolumn{1}{c}{0.476 } & 0.677  & 0.628  \\
    \textbf{BTF} & 0.485  & 0.516  & 0.511  & 0.592  & 0.511  & 0.511  & 0.570  & 0.416  & 0.333  & 0.347  & 0.515  & \multicolumn{1}{c}{0.452 } & 0.589  & 0.461  \\
    \textbf{Patch3D\_40} & 0.557  & 0.829  & 0.722  & 0.690  & 0.844  & 0.830  & 0.767  & 0.831  & 0.865  & 0.777  & 0.417  & \multicolumn{1}{c}{0.683 } & 0.826  & 0.804  \\
\hline
    \end{tabular}%
}
\caption{Simple One-Shot comparison of Patch3D's undershooting on Anomaly-ShapeNet with other approaches P-AUROC results. All experiments were done on two RTX 2080Ti.\label{13}}
\end{table*}

\begin{table*}[htbp]
\centering
\resizebox{\textwidth}{!}{
\begin{tabular}{c|cccccccccccc|cc}
    \hline
    \multicolumn{15}{c}{\textbf{O-AUROC}} \\
    \hline
    \textbf{Method} & \textbf{ashtray0} & \textbf{bag0} & \textbf{bottle0} & \textbf{bottle1} & \textbf{bottle3} & \textbf{bowl0} & \textbf{bowl1} & \textbf{bowl2} & \textbf{bowl3} & \textbf{bowl4} & \textbf{bowl5} & \multicolumn{1}{c}{\textbf{bucket0}} & \textbf{bucket1} & \textbf{cap0} \\
    \hline
    \textbf{Patchcore (FPFH+Raw)} & 0.510  & 0.476  & 0.762  & 0.747  & 0.730  & 0.711  & 0.541  & 0.822  & 0.844  & 0.833  & 0.702  & \multicolumn{1}{c}{0.727 } & 0.613  & 0.841  \\
    \textbf{RegAD} & 0.605  & 0.586  & 0.548  & 0.568  & 0.619  & 0.544  & 0.385  & 0.385  & 0.341  & 0.489  & 0.628  & \multicolumn{1}{c}{0.505 } & 0.371  & 0.581  \\
    \textbf{PatchCore (Raw)} & 0.600  & 0.657  & 0.557  & 0.491  & 0.635  & 0.630  & 0.493  & 0.533  & 0.296  & 0.515  & 0.572  & \multicolumn{1}{c}{0.483 } & 0.371  & 0.537  \\
    \textbf{M3DM} & 0.543  & 0.600  & 0.600  & 0.561  & 0.606  & 0.530  & 0.574  & 0.393  & 0.463  & 0.622  & 0.516  & \multicolumn{1}{c}{0.597 } & 0.413  & 0.470  \\
    \textbf{BTF} & 0.543  & 0.510  & 0.557  & 0.512  & 0.387  & 0.567  & 0.456  & 0.567  & 0.419  & 0.556  & 0.526  & \multicolumn{1}{c}{0.556 } & 0.489  & 0.581  \\
    \textbf{Patch3D\_40} & 0.486  & 0.414  & 0.576  & 0.691  & 0.660  & 0.374  & 0.511  & 0.552  & 0.474  & 0.489  & 0.726  & \multicolumn{1}{c}{0.635 } & 0.406  & 0.659  \\
\hline
    \end{tabular}%

}
\caption{Simple One-Shot comparison of Patch3D's undershooting on Anomaly-ShapeNet with other approaches O-AUROC results. All experiments were done on two RTX 2080Ti.\label{14}}
\end{table*}


\end{document}